%% file: main.tex
\documentclass[]{opendatalab}

\usepackage{fontspec}
\usepackage{xeCJK}
\usepackage{xurl}
\newCJKfontfamily\DianShiCJK[
  Path=fonts/,
  Extension=.otf,
  UprightFont=*,
  ItalicFont=NotoSerifCJKsc-Regular
]{NotoSerifCJKsc-Regular}
\usepackage{xcolor}
\usepackage{longtable}
\usepackage{booktabs}
\usepackage{needspace}
\usepackage{amssymb}
\usepackage{array}
\usepackage{calc}

\usepackage{adjustbox}
\usepackage{pdflscape}
\usepackage{listings}
\usepackage{rotating}
\usepackage{float}
\usepackage[misc]{ifsym}

\newcommand{\name}[1]{\textsf{#1}}

\input{xref}

\title{Reaction precedents extend retrosynthetic prediction beyond training coverage}

\author[1\dag]{Jiaxing Sun}
\author[1\dag]{Xuli Shen}
\author[1\dag\ddagger]{Jiang Wu}
\author[2\dag]{Ruiyi Ding}
\author[1,3\dag]{Ruikang Zhang}
\author[1,4\dag]{Xiao An}
\author[1\dag]{Xibao Li}
\author[1\dag]{Yubin Wang}
\author[1\dag]{Xingjian Wei}

\author[1]{Yinfan Wang}
\author[5]{Yu Liu}
\author[6]{Lingli Ge}

\author[7,1]{Xi Xue}
\author[7]{Xiaojian Wang}

\author[1]{Lijun Wu}
\author[8]{Chang-Yu Hsieh}
\author[1]{Yuqiang Li}

\author[1]{Bowen Zhou}
\author[1]{Dahua Lin}
\author[1\ \textrm{\textmd{\Letter}}]{Conghui He}

\affiliation[1]{Shanghai Artificial Intelligence Laboratory}
\affiliation[2]{Fudan University}
\affiliation[3]{Peking University}
\affiliation[4]{Wuhan University}
\affiliation[5]{University of Science and Technology of China}
\affiliation[6]{Shanghai Jiao Tong University}
\affiliation[7]{Institute of Materia Medica, Chinese Academy of Medical Sciences \& Peking Union Medical College}
\affiliation[8]{Zhejiang University}

\abstract{%
\input{sections/00-abstract}
}

\date{\today}
\metadata[Equal Contribution ($\dagger$)]{Jiaxing Sun, Xuli Shen, Jiang Wu, Ruiyi Ding, Ruikang Zhang, Xiao An, Xibao Li, Yubin Wang, Xingjian Wei}
\metadata[Project Lead ($\ddagger$)]{Jiang Wu, \email{wujiang@pjlab.org.cn}}
\correspondence{Conghui He, \email{heconghui@pjlab.org.cn}}

\begin{document}

\maketitle

\begin{tcolorbox}[colback=odlbg,colframe=odlblue,boxrule=0.5pt,arc=1pt,left=8pt,right=8pt,top=6pt,bottom=6pt]
\small\itshape DianShi, the romanized form of {\DianShiCJK 点石，}evokes the Chinese idiom {\DianShiCJK 点石成金} —literally, “turning stone into gold”—and represents our mission to transform organic-chemistry knowledge scattered across patents, papers, and other literature into structured, AI-ready data and intelligence.
\end{tcolorbox}

\input{sections/01-introduction}
\input{sections/02-results}
\input{sections/03-discussion}
\input{sections/04-methods}
\input{sections/05-data-availability}
\input{sections/06-code-availability}
\input{sections/07-acknowledgements}

\clearpage
\bibliographystyle{unsrtnat}
\bibliography{paper}

\clearpage
\input{sections/11-extended-data}

\clearpage
\input{sections/12-supplementary-information}

\end{document}

%% file: xref.tex
\newcounter{xrfig}              
\newcounter{xredfig}            
\newcounter{xrnote}             
\newcounter{xrnotesec}[xrnote]  
\newcounter{xrtable}            

\renewcommand{\thexrnotesec}{\thexrnote.\arabic{xrnotesec}}

\crefformat{xrfig}{#2Fig. #1#3}
\crefformat{xredfig}{#2Extended Data Fig. #1#3}
\crefformat{xrnote}{#2Supplementary Note #1#3}
\crefformat{xrtable}{#2Supplementary Table #1#3}
\crefformat{xrnotesec}{#2Section #1#3}

\crefmultiformat{xrnote}%
  {#2Supplementary Notes #1#3}{ and #2#1#3}{, #2#1#3}{ and #2#1#3}
\crefmultiformat{xrtable}%
  {#2Supplementary Tables #1#3}{ and #2#1#3}{, #2#1#3}{ and #2#1#3}
\crefmultiformat{xrnotesec}%
  {#2Sections #1#3}{ and #2#1#3}{, #2#1#3}{ and #2#1#3}

\newcommand{\figref}[2][]{\hyperref[fig:#2]{\cref*{fig:#2}#1}}
\newcommand{\edfigref}[1]{\hyperref[edfig:#1]{\cref*{edfig:#1}}}
\newcommand{\noteref}[1]{\hyperref[note:#1]{\cref*{note:#1}}}
\newcommand{\tabref}[1]{\hyperref[tab:#1]{\cref*{tab:#1}}}
\newcommand{\notesec}[1]{\hyperref[notesec:#1]{\cref*{notesec:#1}}}
\newcommand{\secref}[1]{\nameref{#1}}

\makeatletter

\newcommand{\xr@next}[1]{\the\numexpr\value{#1}+1\relax}

\newcommand{\mainfighead}[2]{%
  \refstepcounter{xrfig}\label{fig:#1}%
  \noindent\textbf{Fig.~\thexrfig\ \textbar{} #2}}

\newcommand{\sitablehead}[2]{%
  \refstepcounter{xrtable}\label{tab:#1}%
  \noindent\textbf{Supplementary Table~\thexrtable\ \textbar{} #2}}

\newcommand{\edfighead}[2]{%
  \edef\xr@num{\xr@next{xredfig}}%
  \subsection{Extended Data Fig. \xr@num\ \textbar{} #2}%
  \refstepcounter{xredfig}\label{edfig:#1}}

\newcommand{\sinotehead}[2]{%
  \edef\xr@num{\xr@next{xrnote}}%
  \subsection{Supplementary Note \xr@num. #2}%
  \refstepcounter{xrnote}\label{note:#1}}

\newcommand{\sinotesub}[2]{%
  \refstepcounter{xrnotesec}%
  \subsubsection*{\thexrnotesec\ #2}\label{notesec:#1}}

\makeatother

%% file: sections/00-abstract.tex
Reaction precedents guide retrosynthesis, but how far they can compensate
for gaps in task-specific training coverage remains unclear.
Using 2.71 million reactions, we restricted training and retrieval coverage
by transformation or time, then expanded only the retrieval corpus with
model parameters fixed.
For transformations excluded from training, mean Top-1 accuracy increased
from 13.0\% to 58.0\% on a historical test set and from 4.9\% to 23.1\% on a
temporally held-out test set.
For temporal coverage gaps, retrieval expansion enabled a model trained on
records before 2017 to match or exceed a non-retrieval model trained on
records before 2025 on both test sets.
Without task-specific fine-tuning, precedents improved all 14
representative general-purpose models on both test sets and
\name{USPTO-50K}; the strongest with precedents surpassed all eight
representative specialized models.
These findings show that inference-time reaction precedents can extend
retrosynthetic prediction beyond task-specific training coverage without
parameter updates.

%% file: sections/01-introduction.tex
\section{Introduction}\label{introduction}

Scientific prediction draws on both established knowledge and newly
available evidence.
As AI models are increasingly used for scientific prediction,
they should be able to draw on this growing evidence without being
retrained for every update\citep{lewis2020rag}.
When planning syntheses, chemists complement their acquired expertise by
consulting reaction precedents in the literature and databases.
This motivates a broader question for scientific AI: to what extent can a
trained model expand its predictive scope by drawing on external evidence?

\begin{figure}[tp]
{\centering
\includegraphics[width=0.96\linewidth]{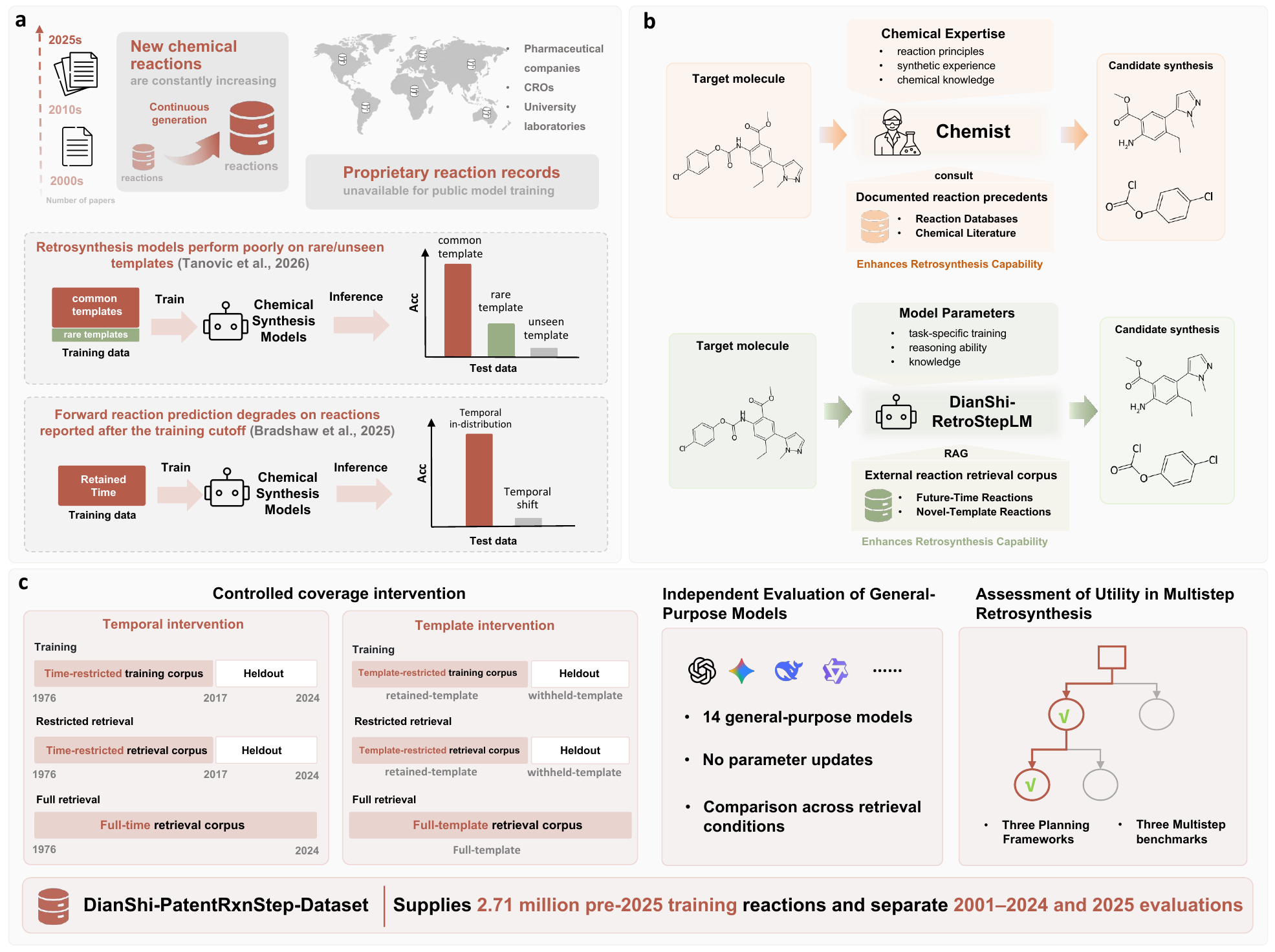}\par}
\vspace{6pt}
{\footnotesize
\mainfighead{1}{Coverage gaps in retrosynthetic prediction and the design of the retrieval-expansion study}.
\textbf{a}, New reactions are continually discovered and reported, so the
record available keeps growing while a trained model's coverage stays frozen
at its training corpus.
A second, independent gap is access: a substantial share of reaction records,
held by pharmaceutical companies and contract research organizations, is
proprietary and never enters public corpora.
Both gaps limit training coverage, and coverage limits prediction: single-step
retrosynthesis models are substantially less accurate on rare or absent
templates, including models that do not explicitly use
templates\citep{tanovic2026bias}, and forward reaction prediction degrades on
reactions from later periods excluded from
training\citep{bradshaw2025generalize}.
Whether reaction precedents supplied at inference can extend predictive
coverage beyond task-specific training without further parameter updates
motivates this study.
\textbf{b}, Reaction knowledge reaches chemists and models in the same two
ways.
A chemist identifies candidate precursors for a target molecule by combining
chemical expertise (reaction principles, synthetic experience and chemical
knowledge) with documented reaction precedents drawn from databases and the
literature.
Likewise, a model can absorb such knowledge into task-specific parameters
during training, where it acts together with the model's reasoning ability and
stored knowledge, or receive it at inference time as retrieved reaction
precedents, which can cover reactions recorded after training or following
templates absent from the training corpus.
\textbf{c}, Study design.
Two complementary interventions restrict coverage in both the training and the
matching retrieval corpora: a temporal intervention retaining only records
before 2017 (\name{Pre2017}; 1976--2016 in training), and a template
intervention withholding about half of the distinct reaction templates
(retained- and withheld-template partitions, S42--S44).
Each intervention pairs a restricted retrieval corpus with a full one
(full-time, 1976--2024; full-template), while the training corpus remains
restricted throughout and evaluation targets are held out.
After training, each fixed checkpoint is re-evaluated on the same targets with
only the retrieval corpus replaced, keeping model parameters, retrieval
procedure and evaluation targets unchanged.
Independently, 14 representative general-purpose models, without parameter
updates, are compared across a no-precedent condition and three retrieval
corpora, and DianShi single-step models are assessed within three planning
frameworks (\name{RetroAgent}, \name{Retro-R1-modified},
\name{DianShi-RetroPlanSystem-v0}) on three multistep benchmarks
(\name{Pistachio Hard}, \name{USPTO-190}, \name{PatMol-150}).
\name{DianShi-PatentRxnStep-Dataset} supplies 2.71 million pre-2025 training
reactions, with separate 2001--2024 and 2025 evaluation sets.
}
\end{figure}

Retrosynthesis---the task of identifying possible precursors to a target
molecule---is central to synthesis planning for medicines and materials.
Models trained on reaction data can support
the search for multistep synthetic
routes\citep{segler2018planning} by proposing precursor candidates for each
target molecule\citep{xiong2025reactseq,xu2025rxngraphormer}.
Their predictive scope, however, depends on the chemistry represented in
training.
Evaluations of single-step retrosynthesis models have shown substantially
lower accuracy for reactions whose templates are rare or absent in the
training data, including in models that do not explicitly use
templates\citep{tanovic2026bias}.
Complementary studies of forward reaction prediction have revealed
limitations when models are evaluated on reactions from later periods
excluded from training\citep{bradshaw2025generalize}.
These coverage gaps have practical relevance: researchers worldwide
continually generate new experimental reaction data after a model has been
trained, while pharmaceutical companies and contract research organizations
(CROs) hold proprietary reaction records that are generally unavailable for public
model training.
Retrosynthesis therefore provides a concrete setting for scientific AI in which
to test whether external evidence can extend a model's predictive scope:
training coverage and the evidence supplied at inference can be varied
independently, with model parameters held fixed.

Previous studies have demonstrated the value of retrieved reaction
information for retrosynthetic
prediction\citep{coley2017retrosim,qian2023textreact,fang2023retrosub,qiao2025rarb,xie2023retroknn}.
Reaction examples have also improved predictions by general-purpose
language models without task-specific
fine-tuning\citep{guo2023chemllmbench,hassen2025atomanchored}.
However, how far reaction precedents can mitigate the predictive limitations
associated with gaps in training coverage remains unclear.
A complementary question concerns the capabilities of general-purpose
models: earlier evaluations reported limited retrosynthetic
performance\citep{guo2023chemllmbench}, but advances in foundation models
motivate reassessing these capabilities and the additional benefit provided
by reaction precedents.

To address these questions, we combined controlled retrieval corpus
expansion experiments with evaluations of contemporary general-purpose
language models.
We constructed \name{DianShi-PatentRxnStep-Dataset}, comprising 2.71
million training reactions recorded before 2025, with separate
evaluation sets for 2001--2024 and 2025.
To assess compensation for gaps in task-specific training coverage, we
restricted transformation or temporal coverage in both the training corpus
and the retrieval corpus.
After training, we held model parameters and the
retrieval procedure fixed and expanded only the retrieval corpus,
comparing predictions on the same targets before and after expansion.
We further evaluated 14 representative general-purpose models with and
without retrieved reaction precedents, without task-specific fine-tuning,
and compared their performance with eight representative released
specialized retrosynthesis models (\figref{1}).

With model parameters held fixed, expanding the retrieval corpus improved
prediction for transformations
absent from the model's fine-tuning data, including those represented by few
records in the expanded corpus.
It also enabled a model trained on a temporally restricted corpus to match
or exceed a non-retrieval model trained on the full corpus, without
further parameter updates.
Without task-specific fine-tuning, retrieved precedents improved all 14
representative general-purpose models across three test sets, and the
strongest general-purpose model with DianShi precedents outperformed all
eight representative released specialized models evaluated here.
The DianShi single-step models also improved route finding across nine
planning settings.
Together, these findings show that inference-time access to reaction
precedents can extend retrosynthetic prediction beyond task-specific
training coverage without further parameter updates.
More broadly, they suggest a way for scientific predictors to benefit from
an expanding body of external evidence without retraining for every update.

%% file: sections/02-results.tex
\section{Results}\label{results}

\subsection{A common data foundation for controlled retrieval corpus
expansion}\label{a-common-data-foundation-for-controlled-retrieval-corpus-expansion}

We constructed \name{DianShi-PatentRxnStep-Dataset} from \name{DianShi-RxnDB}\citep{wang2026dianshirxndb}, a large-scale, provenance-linked organic reaction-data platform that we previously developed, to study
reaction evidence accessed through task-specific training and
inference-time retrieval.
Following standardization, deduplication and
reaction-quality filtering, the dataset comprised 2,712,259 training
reactions recorded before 2025 (\name{Train-Full}), together with separate
validation and test sets.
To assess prediction on historical chemistry and under temporal shift, we
defined two validation--test pairs: \name{Val-2001-2024} and
\name{Test-2001-2024}, each containing 3,798 reactions from 2001 to 2024;
and \name{Val-2025} and \name{Test-2025}, each containing
1,886 reactions from 2025 and therefore
temporally held out from pre-2025 training.
The five training, validation and test partitions were disjoint
by standardized target product, with additional product-level exclusions
against external comparison datasets (\secref{methods}; \tabref{1}).
\name{Train-Full} also served as the full retrieval corpus (\cref{note:1,note:2}).

We developed two single-step predictors from the \name{Qwen3-4B-Base} model\citep{yang2025qwen3}.
\name{DianShi-RetroStepLM-Direct} predicted precursor sets from the target
product alone, whereas \name{DianShi-RetroStepLM-RAG}, a retrieval-augmented
generation (RAG) model, additionally received five retrieved reaction precedents.
\name{RetroStepLM-RAG} was trained by retrieval-conditioned supervised fine-tuning, in which the target product and its five precedents were supplied together as input, and received precedents in the same format at inference.
We refer to these two models hereafter as \name{RetroStepLM-Direct} and \name{RetroStepLM-RAG}.

To determine whether inference-time evidence could compensate for gaps in
task-specific training coverage, we constructed complementary template-
and time-restricted training corpora, each with a matching restricted
retrieval corpus.
The template intervention removed all reactions bearing selected
templates from both the task-specific training corpus and the retrieval
corpus; the temporal intervention retained only records
before 2017 (\name{Pre2017}).
After training each \name{RetroStepLM-RAG} model, we froze its checkpoint and replaced
only its restricted retrieval corpus with the full retrieval corpus, keeping
all other components and the evaluation targets unchanged.
Accuracy changes were then computed on retained-
and withheld-template reactions in the template intervention and on
\name{Test-2001-2024} and \name{Test-2025} in the temporal intervention.
This design
isolated the effect of expanding accessible reaction evidence without
further parameter updates (\figref{1}; \cref{note:1,note:4}).

\subsection{Expanding the retrieval corpus extends predictive coverage
without further training}\label{expanding-the-retrieval-corpus-extends-predictive-coverage-without-further-training}

\begin{figure}[t]
{\centering
\includegraphics[width=\linewidth]{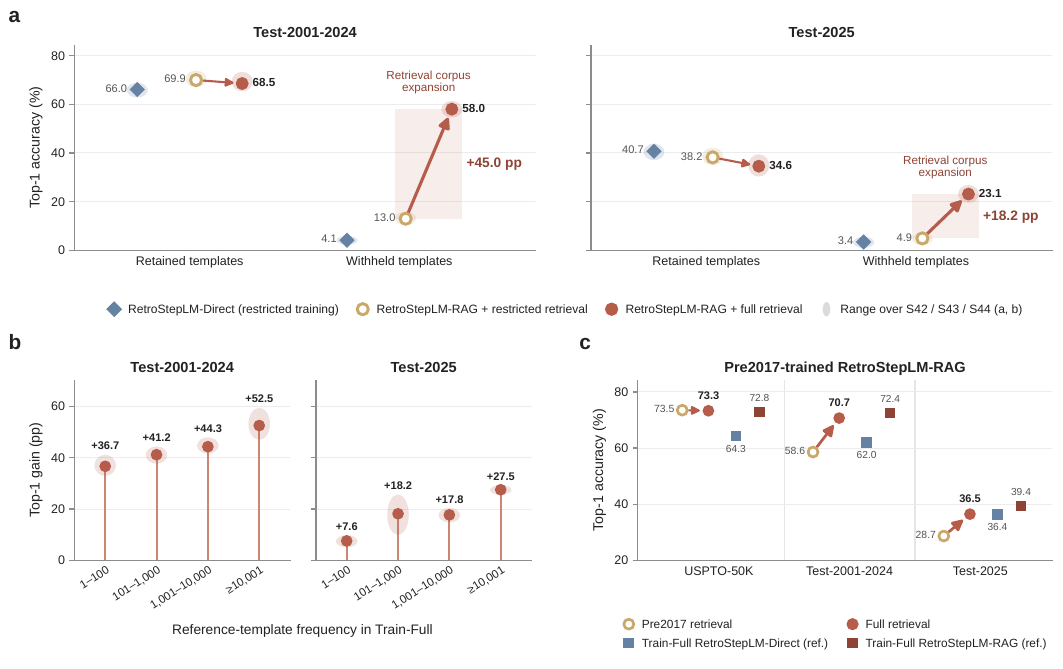}\par}
\vspace{6pt}
{\footnotesize
\mainfighead{2}{Expanding the retrieval corpus extends predictive coverage without further training}.
\textbf{a}, Top-1 accuracy (exact match of the complete predicted precursor set
against the recorded reference) on retained- and withheld-template reactions in
\name{Test-2001-2024} (left) and \name{Test-2025} (right).
All points are arithmetic means over the three template partitions
(S42, S43, S44), with light shading spanning the partitions and reactions not
pooled across them.
The two \name{RetroStepLM-RAG} conditions share one fixed checkpoint and differ
only in the retrieval corpus; \name{RetroStepLM-Direct} is separately trained.
\textbf{b}, Paired accuracy gain (full minus restricted retrieval; percentage
points) on withheld-template reactions across four reference-template
frequency ranges in \name{Train-Full} (1--100, 101--1,000, 1,001--10,000 and
$\geq$10,001 records).
\textbf{c}, Top-1 accuracy on \name{USPTO-50K}, \name{Test-2001-2024} and
\name{Test-2025} (left to right) for a fixed \name{Pre2017-trained}
\name{RetroStepLM-RAG} checkpoint with \name{Pre2017} or full retrieval,
together with \name{Train-Full}-trained \name{RetroStepLM-Direct} and
\name{RetroStepLM-RAG} as separately trained references.
Results and denominators are reported in \cref{note:1,note:4}.
}
\end{figure}

We first investigated whether expanding the retrieval corpus could improve
prediction for transformations absent from the model's fine-tuning data.
We constructed three alternative template partitions using partition
seeds 42, 43 and 44, representing different withheld-template assignments
rather than independent training replicates.
In each partition, approximately half of the distinct local reaction
templates represented in \name{Test-2001-2024} and \name{Test-2025} were
withheld, and all training reactions bearing a withheld template were
removed from both the task-specific training corpus and the retrieval
corpus (Methods).
The
resulting restricted corpora were denoted \name{TemplateRestricted-S42},
\name{TemplateRestricted-S43} and \name{TemplateRestricted-S44}.
With each trained
\name{RetroStepLM-RAG} checkpoint fixed, replacing only its matching restricted
retrieval corpus with the full pre-2025 retrieval corpus improved Top-1
accuracy on withheld-template reactions in every partition (\figref[a]{2};
\edfigref{1}; \noteref{4}).
The arithmetic mean of the
three partition-level accuracies rose from 13.0\% to 58.0\% in
\name{Test-2001-2024} and from 4.9\% to 23.1\% in \name{Test-2025}.
The proportional
increase was similar in \name{Test-2001-2024} and \name{Test-2025} (approximately
4.5-fold versus 4.7-fold; partition-level folds of 4.0--4.8 and 4.1--5.7,
respectively).
In the same paired comparisons, the
corresponding mean on retained-template reactions decreased from 69.9\%
to 68.5\% in \name{Test-2001-2024} and from 38.2\% to 34.6\% in \name{Test-2025}, with
paired decreases in every partition of 0.6--2.1 and 2.7--4.7 percentage
points, respectively.
These decreases were modest relative to the gains on
withheld-template reactions.
Retrieval corpus expansion therefore substantially
improved prediction where task-specific training lacked transformation
coverage, with similar proportional improvement on historical chemistry
in \name{Test-2001-2024} and temporally held-out chemistry in \name{Test-2025}.

These benefits were not confined to transformations with abundant
precedents.
We stratified withheld-template reactions by the frequency of
their local reaction template in the full pre-2025 retrieval
corpus (1--100, 101--1,000, 1,001--10,000 and $\geq$10,001) and found
that corpus expansion improved Top-1 accuracy in all four ranges in both
\name{Test-2001-2024} and \name{Test-2025} (\figref[b]{2}; \edfigref{2};
\noteref{4}).
For templates represented by 1--100 reactions in the
full corpus, mean partition-level gains were 36.7 percentage points in
\name{Test-2001-2024} and 7.6 percentage points in \name{Test-2025} (ranges across
partitions: 34.4--39.7 and 6.6--8.4).
These results indicate that
corpus expansion can benefit transformations with relatively few
recorded precedents.

We then examined the complementary temporal intervention: whether
access to reaction records beyond the training period could improve
prediction without further training.
\name{RetroStepLM-RAG} was trained on the \name{Pre2017} subset
(1.72 million reactions; 63.45\% of \name{Train-Full}), with precedents
retrieved from the same corpus.
After training, we fixed the checkpoint and
expanded only the retrieval corpus to include the remaining 2017--2024
records in \name{Train-Full}.
This increased Top-1 accuracy from 58.6\% to
70.7\% on \name{Test-2001-2024} (+12.1 percentage points) and from 28.7\% to
36.5\% on \name{Test-2025} (+7.8 percentage points) (\figref[c]{2};
\edfigref{3}; \noteref{4}).

With the expanded retrieval corpus, the \name{Pre2017}-trained model exceeded
full-corpus \name{RetroStepLM-Direct} on \name{Test-2001-2024} (70.7\% versus 62.0\%) and matched its
accuracy on \name{Test-2025} (36.5\% versus 36.4\%), despite having
been trained on only 1.72 million of the 2.71 million reactions.
Inference-time
access to the additional records thus compensated sufficiently for the
temporal training restriction to match or exceed full-corpus \name{RetroStepLM-Direct}
on both test sets.
Even so, accuracy remained below that of \name{RetroStepLM-RAG} trained
on \name{Train-Full} and evaluated with the full retrieval corpus: 70.7\%
versus 72.4\% on \name{Test-2001-2024} and 36.5\% versus 39.4\% on \name{Test-2025}.

On \name{USPTO-50K}\citep{coley2017retrosim}, which comprises reactions from patents
published between 1976 and 2016, retrieval corpus expansion left accuracy
essentially unchanged (73.5\% versus 73.3\%).
This limited benefit may
reflect the benchmark's temporal coverage, which falls entirely within
the \name{Pre2017} training period, whereas expansion added the
2017--2024 records.
\subsection{Reaction precedents improve retrosynthesis without
task-specific training}\label{reaction-precedents-improve-retrosynthesis-without-task-specific-training}

\begin{figure}[t]
{\centering
\includegraphics[width=\linewidth]{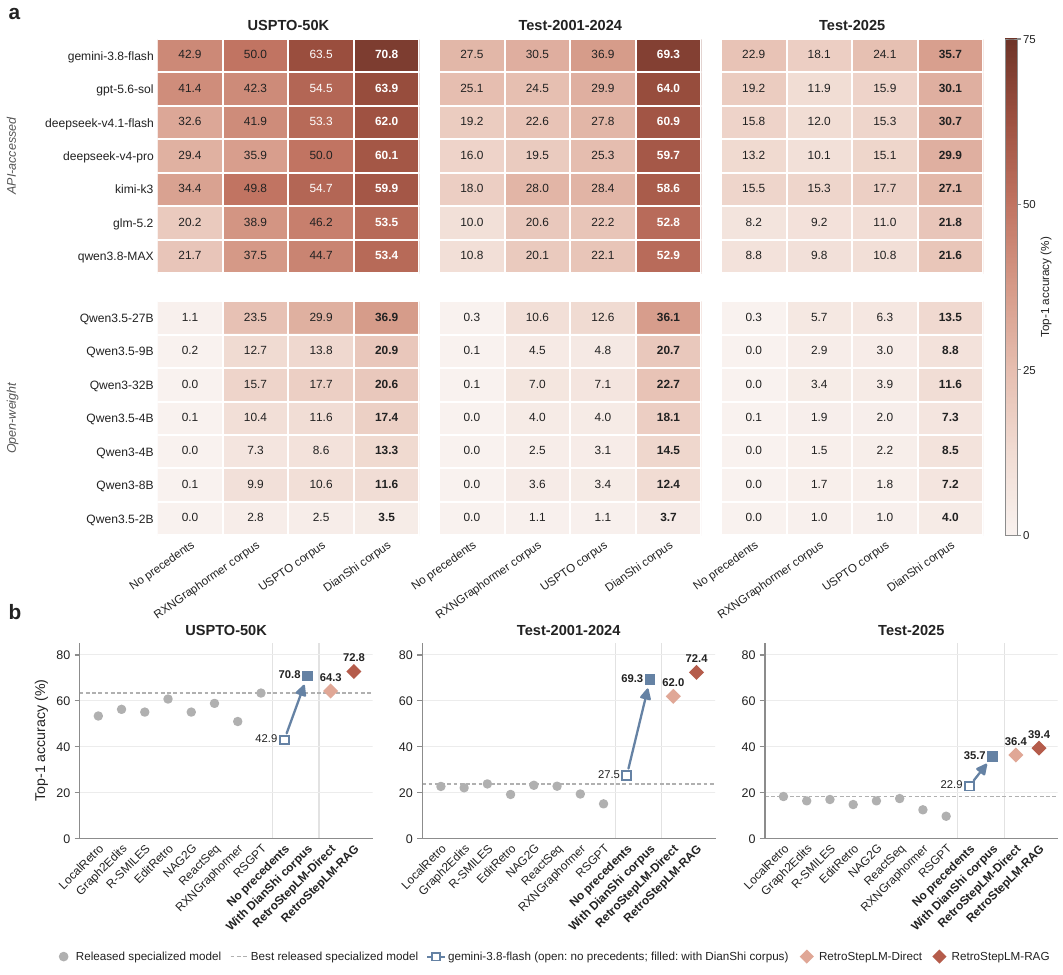}\par}
\vspace{6pt}
{\footnotesize
\mainfighead{3}{Reaction precedents improve retrosynthesis without task-specific training}.
\textbf{a}, Top-1 accuracy of 14 representative general-purpose models on
\name{USPTO-50K}, \name{Test-2001-2024} and \name{Test-2025} (left to right).
Within each test set, columns compare no precedents with five precedents
retrieved from the \name{RXNGraphormer}, USPTO or DianShi corpus.
The DianShi corpus is derived from \name{Train-Full}, from which products in
the DianShi evaluation partitions are excluded; the USPTO corpus merges the
training sets of \name{USPTO-50K}, \name{USPTO-MIT} and \name{USPTO-Full}; the
\name{RXNGraphormer} corpus is constructed from the reactions used to pretrain
\name{RXNGraphormer}.
\textbf{b}, Accuracy of eight representative released specialized
models alongside \name{gemini-3.8-flash} without precedents (open symbol) or
with the DianShi corpus (filled symbol), and of \name{RetroStepLM-Direct} and
\name{RetroStepLM-RAG} trained on \name{Train-Full}, on \name{USPTO-50K},
\name{Test-2001-2024} and \name{Test-2025} (left to right).
Results are reported in \noteref{6}.
}
\end{figure}

We evaluated whether general-purpose models could use reaction
precedents without task-specific parameter updates, building on earlier
demonstrations that frozen general-purpose models can benefit from
retrieved or same-class reaction examples\citep{guo2023chemllmbench,hassen2025atomanchored}.
Using a common retrieval-and-prompting framework, we supplied five
precedents for each target product to 14 representative general-purpose
models and compared their predictions with those generated without
retrieved precedents.
To assess how the retrieval corpus affected prediction, precedents were
drawn from three retrieval corpora: the full
DianShi retrieval corpus, supplied by \name{Train-Full}; a USPTO retrieval
corpus assembled by merging the training sets of
\name{USPTO-50K}\citep{coley2017retrosim}, \name{USPTO-MIT}\citep{jin2017wln} and
\name{USPTO-Full}\citep{dai2019gln}; and a retrieval corpus constructed from
the reactions used to pretrain \name{RXNGraphormer}\citep{xu2025rxngraphormer},
referred to hereafter as the \name{RXNGraphormer} corpus (\secref{methods};
\noteref{6}).
Across all 42 model--test-set combinations, Top-1 accuracy was highest
when precedents were retrieved from the DianShi corpus, above the USPTO and
\name{RXNGraphormer} corpora in every case (\figref[a]{3}; \tabref{15}).
Precedents from the external corpora were not uniformly better than none:
on \name{Test-2025}, \name{gemini-3.8-flash} reached 18.12\% with
\name{RXNGraphormer} precedents, below its 22.85\% without retrieved
precedents, whereas DianShi precedents raised accuracy to 35.74\%.

The strongest general-purpose model, \name{gemini-3.8-flash} with
DianShi precedents, achieved Top-1 accuracies of 70.80\%, 69.33\% and
35.74\% on \name{USPTO-50K}, \name{Test-2001-2024} and \name{Test-2025}, respectively,
compared with 42.94\%, 27.49\% and 22.85\% without retrieved precedents.
On each test set, \name{gemini-3.8-flash} with DianShi precedents exceeded
the accuracies of all eight representative released specialized
models evaluated here:
\name{LocalRetro}\citep{chen2021localretro}, \name{Graph2Edits}\citep{zhong2023graph2edits},
\name{R-SMILES}\citep{zhong2022rsmiles}, \name{EditRetro}\citep{han2024editretro},
\name{NAG2G}\citep{yao2024nag2g}, \name{ReactSeq}\citep{xiong2025reactseq},
\name{RXNGraphormer}\citep{xu2025rxngraphormer} and \name{RSGPT}\citep{deng2025rsgpt}
(\figref[b]{3}; \noteref{6}).
The benefit was not confined to large API-accessed models:
\name{Qwen3.5-27B} rose from 1.08\%, 0.26\% and 0.27\% without precedents
to 36.85\%, 36.12\% and 13.52\% with DianShi precedents.
Its accuracy with precedents exceeded that of \name{glm-5.2} without
precedents (20.17\%, 9.98\% and 8.17\%), a 744-billion-parameter sparse
mixture-of-experts model\citep{zai2026glm52}, although with the same
precedents \name{glm-5.2} remained higher (53.45\%, 52.79\% and 21.84\%;
\tabref{15}).
Together, these comparisons show that inference-time access to reaction
precedents enabled general-purpose models to surpass the evaluated
task-specialized predictors without task-specific training.

As a task-specialized reference, the 4B-parameter \name{RetroStepLM-RAG} model trained
on \name{Train-Full} and evaluated with the full retrieval corpus achieved Top-1
accuracies of 72.8\%, 72.4\% and 39.4\% on \name{USPTO-50K}, \name{Test-2001-2024} and
\name{Test-2025}, respectively---the highest among the predictors evaluated
here (\noteref{6}).
It outperformed both full-corpus \name{RetroStepLM-Direct} and
the strongest general-purpose model supplied with DianShi precedents on
all three test sets.

Within a fixed retrieval corpus, the usefulness of precedents also
depended on which reactions were selected for each target.
We used reference reaction templates to constrain precedent
selection in separate diagnostics across four of the 14 representative
general-purpose models and the \name{RetroStepLM-RAG} model trained on \name{Train-Full}.
Matching-template exclusion reduced accuracy on all three test sets
relative to standard retrieval.
On \name{Test-2025}, matching-template exclusion fell below the
no-precedent baseline for
\name{gemini-3.8-flash} (19.35\% versus 22.85\%) and \name{deepseek-v4.1-flash}
(8.43\% versus 15.85\%).
Reference-template restriction improved prediction for all five models
across the three test sets (\edfigref{4}; \noteref{5}).
These results show
that product-structure similarity alone does not ensure useful reaction
precedents, so whether an expanded corpus helps also depends on how
precedents are selected for each target.

Together, these results show that reaction precedents can support strong
retrosynthetic prediction without task-specific training, while the
highest accuracy was achieved by combining retrieval-conditioned fine-tuning
with inference-time access to reaction precedents.
Task-specific training and inference-time evidence thus provided
complementary benefits, with the utility of that evidence depending on
precedent selection.
\subsection{DianShi single-step models improve solve rates across
planning frameworks}\label{dianshi-single-step-models-improve-solve-rates-across-planning-frameworks}

\begin{figure}[t]
{\centering
\includegraphics[width=\linewidth]{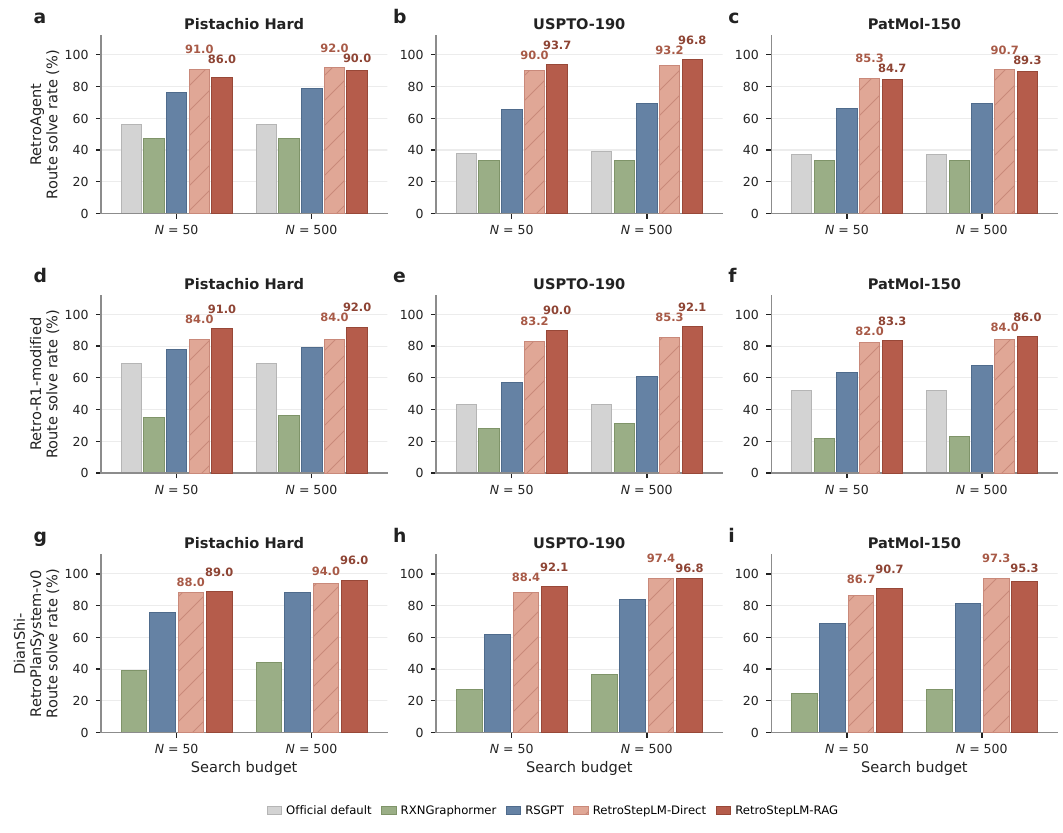}\par}
\vspace{6pt}
{\footnotesize
\mainfighead{4}{DianShi single-step models improve solve rates across planning frameworks}.
\textbf{a--i}, Route solve rates within \name{RetroAgent} (\textbf{a--c}),
\name{Retro-R1-modified} (\textbf{d--f}) and \name{DianShi-RetroPlanSystem-v0}
(\textbf{g--i}); columns show \name{Pistachio Hard} (100 targets),
\name{USPTO-190} (190) and \name{PatMol-150} (150), respectively.
Solve rate counts targets for which at least one complete route to the
inventory was found.
Within each framework, planner, starting-material inventory and search
protocol are fixed; the complete budget trajectories are shown in
\edfigref{5}.
Protocols and complete results are reported in \noteref{7}.
}
\end{figure}

We next evaluated whether the DianShi single-step models improve
downstream route finding.
We used three planning frameworks:
\name{RetroAgent}\citep{zhu2026retroagent}, an LLM-based agentic planner;
\name{Retro-R1-modified}\citep{liu2025retror1}, an adapted \name{Retro-R1} pipeline;
and our own planning framework, \name{DianShi-RetroPlanSystem-v0}.
Each
framework was evaluated on three benchmarks: Pistachio
Hard\citep{yu2024doubleended} (100 targets; evaluated here without the
original specified-starting-material constraint),
\name{USPTO-190}\citep{chen2020retrostar} (190 targets) and \name{PatMol-150}, a
benchmark we constructed from 2025 public patents (150 targets, reference
routes of 7--13 steps).
Within each framework, we compared single-step
models under a fixed planner, starting-material inventory and search
protocol, with all models using the same template-based reaction
validation\citep{song2025aotstar} and their respective template libraries.
We measured solve
rate as the fraction of targets for which a complete route to the
starting-material inventory was found within the search budget (\figref{4};
\edfigref{5}; \noteref{7}).

Both DianShi models achieved higher solve rates than every evaluated
non-DianShi model across all nine planner--benchmark settings, at both
N=50 and N=500.
At N=500, their solve rates ranged from 84.00\% to
97.37\%, exceeding the strongest non-DianShi model in each setting by
5.00--31.06 percentage points.

\name{RetroStepLM-RAG} performed better overall in multistep planning, achieving
higher solve rates than \name{RetroStepLM-Direct} in seven of the nine settings at N=50 and in five of the nine at N=500, with
advantages of 1.00--7.00 and 2.00--8.00 percentage points, respectively.
\name{RetroStepLM-Direct} achieved higher
solve rates in the remaining two and four settings, with advantages of
0.66--5.00 and 0.53--2.00 percentage points.
These budget-dependent differences are consistent with the higher single-step Top-1
accuracy of \name{RetroStepLM-RAG} on all three test sets, but they also indicate that
candidate-set composition and search dynamics influence multistep
performance\citep{torrenperaire2024modelsmatter,choe2025crosstalk}.
Together, these results show that stronger single-step models can
substantially improve multistep planning, while single-step accuracy
alone does not determine multistep utility.

%% file: sections/03-discussion.tex
\section{Discussion}\label{discussion}

Our controlled interventions with a language model trained through retrieval-conditioned fine-tuning show that
retrosynthetic prediction can extend beyond task-specific training
coverage.
With model parameters held fixed,
expanding the inference-time retrieval corpus improved
prediction for transformations absent from the model's fine-tuning data.
One
interpretation is that retrieval-conditioned fine-tuning equips the model not only
with knowledge of particular transformations, but also with an ability
to use reaction precedents supplied at inference.
External evidence can
then provide information needed for prediction without requiring each
additional transformation to be incorporated through further parameter
training\citep{lewis2020rag}.
Benefits extended to transformations with
few recorded precedents and to records beyond the training period.
These two gaps parallel the shift types that prior work constructs as separate out-of-distribution splits, in which either transformation types or target molecules are held out\citep{yu2024retroood}; here the restrictions acted on corpus coverage rather than on evaluation splits, so that the recovered accuracy could be attributed to retrieved evidence alone.
Inference-time access
to external evidence therefore provides a means of extending the useful
scope of a trained predictor.
This offers a route for incorporating newly reported reactions and proprietary records held by pharmaceutical companies and CROs into prediction without updating model parameters.

The extent of these gains, however, also depended on which precedents were
selected for each target.
Structurally similar products can arise from different
transformations, so product similarity does not guarantee that the selected precedents share the transformation of the recorded reaction.
Diagnostics that use template information from the recorded answer isolate this dependence: restricting precedent selection
to matching-template reactions increased reference-set accuracy on all three test sets, whereas
excluding such reactions reduced it in every model and test set, and on \name{Test-2025} fell below the no-precedent reference for two general-purpose models.
Precedent selection therefore influences agreement with the recorded reaction even when model parameters and the retrieval corpus remain fixed.
The expansion did not raise accuracy uniformly either: withheld-template reactions improved substantially, whereas retained-template reactions, whose transformations were already represented in the fine-tuning data, decreased modestly.
Whether an expanded corpus improves accuracy therefore depends on whether the additional records contain the transformation a target requires, and on how those records enter the precedent set for that target.

The general-purpose model results show that contemporary foundation
models can achieve strong retrosynthetic performance with reaction
precedents and without additional task-specific fine-tuning.
The
strongest general-purpose model with DianShi precedents outperformed all
eight representative released specialized models evaluated across
the three test sets.
These
findings suggest that leading foundation models possess substantial
chemical knowledge and reasoning capabilities that can be brought to bear
on retrosynthesis through access to relevant reaction examples.
Performance also depended on the corpus from which precedents were
retrieved, highlighting the importance of external evidence alongside
model capability.
Related work has advanced retrosynthesis mainly by internalizing reaction knowledge in model parameters---through domain-specific pretraining and supervised training on reaction data\citep{xiong2025reactseq,xu2025rxngraphormer,deng2025rsgpt,zhang2025llmorganic}, or through reinforcement-learning and reasoning-oriented post-training\citep{zhang2025retrodfm,ko2026retroreasoner,li2026retroexpert}.
Extending either route requires further parameter training.
Reaction precedents supplied at inference, by contrast, extended coverage in the present experiments with the checkpoint unchanged, and can complement both approaches for chemical language models.
At the same time, the strongest predictor evaluated here combined
retrieval-conditioned fine-tuning with inference-time access to reaction precedents, supporting
the complementary roles of parameter adaptation and inference-time
evidence.
This combination parallels how chemists draw on acquired expertise and
consult the literature and reaction databases for precedents relevant
to a target molecule.
External reaction evidence can therefore complement learned capabilities without
requiring every additional reaction to be encoded in model parameters.

The planning results demonstrate the practical value of improved
single-step models for multistep synthesis planning.
Both DianShi models
increased solve rates across all nine planner--benchmark settings,
showing that their downstream benefits extended across different planning
frameworks and target collections.
These consistent gains underscore the
importance of single-step candidate generation to route finding: the
reactions proposed at each expansion determine which synthetic pathways
the planner can explore.
These gains were not fully captured by
single-step Top-1 accuracy.
\name{RetroStepLM-RAG} performed better overall in multistep
planning, consistent with its higher single-step accuracy, but its
advantage did not extend to every planning setting.
This variation
suggests that the utility of candidate reactions during search matters
alongside agreement with recorded precursors\citep{torrenperaire2024modelsmatter,choe2025crosstalk}.
Exact matching against one recorded precursor set does not capture all
chemically plausible alternatives\citep{maziarz2026retrochimera,zagribelnyy2026singleanswer}.
Ultimately, computational route finding must be complemented by
experimental validation of the proposed reactions and their conditions\citep{mikulakklucznik2020chematika}.

%% file: sections/04-methods.tex
\section{Methods}\label{methods}

\subsection{Reaction dataset construction and
partitioning}\label{reaction-dataset-construction-and-partitioning}

\name{DianShi-PatentRxnStep-Dataset} was derived from \name{DianShi-RxnDB}, a
provenance-linked organic reaction-data platform that we previously
developed\citep{wang2026dianshirxndb}.
\name{DianShi-RxnDB} uses a fully automated pipeline to extract structured
reaction records from patent text, images and reaction schemes, primarily
from USPTO and EPO organic-synthesis patents published between 1976 and
2025.
Each record represents a reported single-step experiment and links
to its source patent.
We curated these records for single-step
retrosynthesis through reaction standardization, deduplication, quality
filtering and product-disjoint partitioning (\noteref{2}).

Reactant and product SMILES were standardized using RDKit\citep{rdkit}.
Reaction-level duplicates were identified using a standardized
reactant-set--product-set signature, with atom-mapping numbers removed
from the comparison representation and molecular components sorted
deterministically.
For duplicate records spanning multiple years, the
record with the earliest year recorded in \name{DianShi-RxnDB} was retained.
Quality filtering followed the curation conventions used for single-step
retrosynthesis datasets\citep{tanovic2026bias} and excluded reactions with
multiple products, repeated molecules on either side, structural overlap
between reactants and products, more than four reactant components,
inorganic products, inconsistent atom mappings or unsuccessful
reaction-centre or template extraction.
Local reaction templates\citep{chen2021localretro} were grouped across the
quality-controlled corpus, and templates represented by fewer than six
reactions were removed before temporal partitioning, following the cutoff
used in previous template-frequency analyses\citep{tanovic2026bias}.
Detailed filtering and partitioning procedures are provided in
\noteref{2}.

For temporal evaluation\citep{bradshaw2025generalize,qian2023textreact}, partitions were defined using the year of the source patent recorded in
\name{DianShi-RxnDB}.
Records before 2025 supplied the
training and historical evaluation pools, whereas records from
2025 supplied the temporally held-out evaluation pool.
\name{Val-2001-2024}
and \name{Test-2001-2024} were sampled from 2001--2024 records with
stratification by year and template frequency in the full training
corpus; \name{Val-2025} and \name{Test-2025} were sampled from the 2025 pool with
stratification by patent type and template frequency.
The five
partitions---\name{Train-Full}, \name{Val-2001-2024}, \name{Test-2001-2024}, \name{Val-2025} and
\name{Test-2025}---were constructed jointly and were mutually disjoint by
standardized target product\citep{tanovic2026bias}.
To control overlap
with external benchmarks, training candidates were screened against
products in the validation and test partitions of
\name{USPTO-50K}\citep{coley2017retrosim}, \name{USPTO-Full}\citep{dai2019gln} and \name{USPTO-MIT}\citep{jin2017wln}.
DianShi validation and test candidates were additionally screened against
products in the training partitions of these USPTO datasets and of the
corpus used to pretrain \name{RXNGraphormer}\citep{xu2025rxngraphormer}.

The resulting dataset comprised 2,712,259 pre-2025 training reactions in
\name{Train-Full}, 3,798 reactions each in \name{Val-2001-2024} and \name{Test-2001-2024}, and
1,886 reactions each in \name{Val-2025} and \name{Test-2025}.
\name{Train-Full} also served as the full
retrieval corpus.
Dataset identifiers and partition roles are summarized
in \noteref{1}.

\subsection{Fixed-parameter retrieval corpus expansion
experiments}\label{fixed-parameter-retrieval-corpus-expansion-experiments}

Training coverage can be incomplete in two distinct ways: the transformations a target requires may be absent from the fine-tuning data, or the available data may predate the chemistry of interest.
We therefore imposed two complementary coverage restrictions---one on transformations, one on time---and examined whether reaction precedents retrieved at inference time can compensate for these gaps without further parameter updates.

In every experiment, \name{DianShi-RetroStepLM-RAG} was trained on a restricted training corpus, with precedents drawn from the matching restricted retrieval corpus, and the resulting checkpoint was then fixed.
Each evaluation target was scored with either the restricted retrieval corpus or the full retrieval corpus.
The full retrieval corpus comprised the 2,712,259 reactions in \name{Train-Full}, and each restricted retrieval corpus comprised the reactions of its corresponding restricted training corpus.
Paired comparisons therefore switched only between the two retrieval indices,
keeping the checkpoint, retrieval procedure, precedent serialization,
prompting, decoding and scoring unchanged.
Retrieval and training
procedures are described in \secref{reaction-corpora-and-precedent-retrieval}
and \secref{dianshi-single-step-models-and-training}, respectively.

For the template-coverage intervention, we constructed three alternative
template partitions using template-partition seeds 42, 43 and 44.
Within each partition, approximately half of the distinct reaction
templates represented in each of \name{Test-2001-2024} and \name{Test-2025} were designated as
withheld.
Templates shared between the two test sets received a common
retain-or-withhold assignment.
All reactions bearing withheld templates
were removed from \name{Train-Full}, and the remaining records formed the
corresponding \name{TemplateRestricted-S42}, \name{TemplateRestricted-S43} or
\name{TemplateRestricted-S44} corpus, which served as both the restricted
training corpus and the matching restricted retrieval corpus.
Withheld templates
were therefore absent from both restricted resources, whereas retained evaluation
templates had at least one corresponding training reaction.
The
approximately 50\% withholding target referred to distinct templates
rather than reaction counts.
\name{Test-2001-2024} and \name{Test-2025} were each
divided into retained- and withheld-template subsets according to the
corresponding partition.
The three partitions represented alternative withheld-template
assignments rather than independent training replicates.

To assess whether the effects of retrieval corpus expansion extended to
transformations with few recorded precedents, we stratified
withheld-template evaluation reactions by the number of records bearing
their local reaction template\citep{tanovic2026bias,chen2021localretro}
in \name{Train-Full}, which also constituted the full retrieval corpus.
The four frequency ranges were
1--100, 101--1,000, 1,001--10,000 and at least 10,001 reactions per
template.

For the temporal intervention, the \name{Pre2017} training corpus retained
the \name{Train-Full} reactions before 2017.
\name{RetroStepLM-RAG} was
trained on this corpus with precedents retrieved from the matching
\name{Pre2017} retrieval corpus and, after training, evaluated with either the
\name{Pre2017} retrieval corpus or the full retrieval corpus.

\name{RetroStepLM-Direct} models trained on the corresponding restricted corpora and
\name{RetroStepLM-Direct} and \name{RetroStepLM-RAG} models trained on \name{Train-Full} provided reference
configurations.
Configuration correspondence is provided in \noteref{1};
restricted-corpus sizes and template-partition composition are reported
in \noteref{4}.
Accuracy calculation and cross-partition
reporting are described in \secref{single-step-evaluation}.

\subsection{Reaction corpora and precedent
retrieval}\label{reaction-corpora-and-precedent-retrieval}

Retrieval indices were constructed from \name{Train-Full} and each restricted
training corpus defined above, without additional corpus-level
subsampling.
Two further retrieval corpora were constructed for the
general-purpose model evaluations: a USPTO retrieval corpus assembled by
merging the training sets of
\name{USPTO-50K}\citep{coley2017retrosim}, \name{USPTO-MIT}\citep{jin2017wln} and
\name{USPTO-Full}\citep{dai2019gln}, and an \name{RXNGraphormer} retrieval
corpus built from the \name{RXNGraphormer} training corpus\citep{xu2025rxngraphormer}.
Retrieval from all corpora used the same procedure and index
settings, so that differences between corpora reflected their
reaction content rather than the retrieval implementation.
External retrieval
corpora are summarized in \noteref{6}.

Standard retrieval used a two-stage structural-similarity search based on
the target product alone, without reference reaction-class or template
information.
Products were represented by extended-connectivity
fingerprints (ECFPs)\citep{rogers2010ecfp}.
A binary flat index returned
200 candidate reactions ranked by Hamming distance, which were then
reranked by the Tanimoto similarity between their product fingerprints
and the query fingerprint.
Reference-informed modifications to precedent
selection are described in \secref{precedent-relevance-diagnostics}.

During \name{RetroStepLM-RAG} fine-tuning, the query reaction, duplicate reactions and
potentially answer-disclosing records from the same patent were excluded
from candidate precedents.
At each training epoch, five eligible
precedents were randomly selected from the ten highest-similarity
candidates.
At standard inference, \name{RetroStepLM-RAG} and the general-purpose models
received the five highest-ranked eligible precedents deterministically.
Product-level exclusions during dataset construction prevented products
in the DianShi evaluation partitions and the USPTO-series validation and
test partitions from appearing in the DianShi training corpus or its
derived retrieval corpora.

Each precedent contained standardized product and reactant SMILES,
together with atom-mapping-derived synthon SMILES and corresponding
leaving groups.
Five precedents were supplied alongside the target product
as reference context for \name{RetroStepLM-RAG} and the general-purpose models, using the
structured prompt family provided in \noteref{3}.

\subsection{DianShi single-step models and
training}\label{dianshi-single-step-models-and-training}

\name{DianShi-RetroStepLM-Direct} and \name{DianShi-RetroStepLM-RAG} were developed
from \name{Qwen3-4B-Base}\citep{yang2025qwen3} using full-parameter supervised
fine-tuning.
\name{RetroStepLM-Direct} received the target product alone, whereas \name{RetroStepLM-RAG}
underwent retrieval-conditioned supervised fine-tuning, receiving the target product together with five reaction precedents selected using the
retrieval procedure described in \secref{reaction-corpora-and-precedent-retrieval}.
Both models were trained to generate
the same structured target sequence.

Synthon- and leaving-group-based decompositions have also been used in
structured retrosynthesis models\citep{xiong2025reactseq,zhao2025multitask}.
Training used intermediate-label supervision,
with synthon and leaving-group labels derived from atom-mapped reactions.
Synthons represented
product fragments associated with the retrosynthetic disconnection, and
corresponding leaving groups provided information for reconstructing the
reactants.
Target sequences serialized synthon SMILES, leaving groups and
the final standardized reactant set in the order
\texttt{synthons\_smiles}, \texttt{leaving\_groups} and
\texttt{reactants}, with the final reactant SMILES additionally enclosed
in \texttt{<answer>} tags.
For input \(x\) and target sequence \(y=(y_1,\ldots,y_T)\), training
minimized the autoregressive negative log-likelihood over the complete
target sequence,

$$ \mathcal{L}(\theta)=-\sum_{t=1}^{T}\log p_{\theta}(y_t\mid x,y_{<t}), $$

where \(x\) comprised the target product alone for \name{RetroStepLM-Direct} and the
target product together with retrieved precedents for \name{RetroStepLM-RAG}.

Full-corpus models used \name{Train-Full}, whereas coverage-intervention models
used the corresponding \name{Pre2017}, \name{TemplateRestricted-S42},
\name{TemplateRestricted-S43} or \name{TemplateRestricted-S44} training corpus.
For
each \name{RetroStepLM-RAG} model, training precedents were retrieved from the same record
set that supplied its fine-tuning reactions.
SMILES were space-segmented into atom- and bond-level tokens following the
Molecular Transformer tokenization scheme\citep{schwaller2019moleculartransformer}
and encoded using the native Qwen tokenizer without vocabulary expansion.
Models were trained for ten epochs with an effective batch size of 128
and a learning rate of \(1\times10^{-4}\), using cosine decay and linear
warm-up over the first 5\% of training steps.
Training used bfloat16
precision; hardware and distributed-training settings are provided in
\tabref{4} (\noteref{3}).

At inference, both models used beam search to generate ranked
precursor-set candidates for Top-1, Top-3, Top-5 and Top-10 evaluation.
Only the final reactant set was used for precursor-set scoring, as
described in \secref{single-step-evaluation}.
Decoding and
candidate-extraction procedures are described in \noteref{3}.

\subsection{Precedent-relevance
diagnostics}\label{precedent-relevance-diagnostics}

We conducted reference-informed retrieval diagnostics to examine how
precedent selection affected retrosynthetic prediction.
Evaluations
used the \name{RetroStepLM-RAG} model trained on \name{Train-Full} and four of the
14 general-purpose models described below: \name{gemini-3.8-flash},
\name{deepseek-v4.1-flash}, \name{Qwen3-32B} and \name{Qwen3.5-27B}.
All five models were
evaluated on \name{USPTO-50K}, \name{Test-2001-2024} and \name{Test-2025}.
Across
precedent-selection conditions, each model used the full DianShi retrieval
corpus and received five precedents per query.
Model parameters,
evaluation targets, prompt structure, decoding and scoring settings were
held fixed; only the precedent-selection rule changed.

Standard retrieval, which ranks precedents by target-product structural
similarity, was compared with three reference-informed conditions.
Under reference-class restriction, retrieval selected only precedents
sharing the recorded reference reaction class; under reference-template
restriction, it selected only those sharing the reference reaction
template.
Matching-template exclusion removed reactions bearing the reference
template from precedent selection.
In the matching-template exclusion condition, retrieval continued down
the ranked candidate list to collect five eligible precedents; when fewer
than five remained, eligible precedents were cycled to fill the five
context positions.
Predictions without retrieved precedents provided
additional reference conditions for the four general-purpose models.
Performance was assessed at Top-1 using the complete precursor-set
exact-match metric defined in \secref{single-step-evaluation}.

Reaction classes and templates were obtained from the recorded reference
reactions, making these privileged diagnostics rather than deployable
retrieval settings.
Diagnostic conditions and results are provided in
\noteref{5}.

\subsection{General-purpose model evaluation and specialized-model
comparisons}\label{general-purpose-model-evaluation-and-specialized-model-comparisons}

We evaluated 14 representative general-purpose models with and without
retrieved reaction precedents, without task-specific parameter updates,
and compared their performance with DianShi models and eight
representative released specialized models.
These 14 models comprised seven API-accessed models across major
providers and seven open-weight Qwen models spanning 2B to 32B
parameters.
Each model was evaluated on \name{USPTO-50K}, \name{Test-2001-2024} and
\name{Test-2025} under four conditions: no retrieved precedents, or five
precedents retrieved from the \name{RXNGraphormer} retrieval corpus, the USPTO
retrieval corpus or the full DianShi retrieval corpus.
Model identities are provided in
\noteref{6}.

General-purpose models produced one final precursor-set prediction per
target and were assessed at Top-1 using the common scoring procedure
described in \secref{single-step-evaluation}.
API requests used temperature 0, top-p 1.0, a maximum
output-token setting of 65,535 and provider-default thinking settings.
Incomplete API responses were rerun until a complete response was
obtained.
General-purpose model prompting and API request settings are
described in \cref{note:3,note:6}.

For specialized-model comparisons, we evaluated the released weights of
eight representative specialized retrosynthesis models covering the main
published single-step approaches:
\name{LocalRetro}\citep{chen2021localretro}, \name{Graph2Edits}\citep{zhong2023graph2edits},
\name{R-SMILES}\citep{zhong2022rsmiles}, \name{EditRetro}\citep{han2024editretro},
\name{NAG2G}\citep{yao2024nag2g}, \name{ReactSeq}\citep{xiong2025reactseq},
\name{RXNGraphormer}\citep{xu2025rxngraphormer} and \name{RSGPT}\citep{deng2025rsgpt},
without retraining on DianShi data.
Inputs were formatted according to each
method's requirements, and \name{Test-2001-2024} and \name{Test-2025} inference used no
molecular test-time augmentation.
Predictions were assessed using the
common precursor-set matching metric described in \secref{single-step-evaluation}.
Specialized-model evaluation protocols and complete comparison results are
provided in \tabref{17} (\noteref{6}).

\subsection{Single-step evaluation}\label{single-step-evaluation}

Single-step prediction was evaluated by exact matching of complete
precursor sets against the recorded reference.
Final precursor-set
predictions were extracted from each model's output according to its
output format.
Predicted and reference molecules were standardized using
RDKit and converted to canonical SMILES, with molecular components sorted
deterministically before comparison\citep{maziarz2025syntheseus}.
Component order therefore did not affect correctness.
Missing
predictions, invalid molecular strings and partial precursor-set matches
were counted as incorrect.

For specialized models, Top-\(k\) accuracy measured whether at least one
of the first \(k\) ranked candidates matched the complete reference
precursor set.
For \(M\) evaluation reactions, reference set \(R_i\) and
candidate \(\hat{R}_{ij}\) at rank \(j\), the metric was

$$ \mathrm{EM}@k=\frac{1}{M}\sum_{i=1}^{M}\mathbf{1}\!\left[\exists j\leq k:\hat{R}_{ij}=R_i\right],\qquad k\in\{1,3,5,10\}. $$

Candidates were evaluated in their generated rank order.
General-purpose models produced one scored precursor set per target and
were assessed at Top-1.
Accuracy measured agreement with the recorded
reference precursor set.

For template-coverage analyses, cross-partition summaries were arithmetic
means of the three template-partition-specific accuracies, calculated
separately for each test set, retained- or withheld-template subset and,
where applicable, precedent-frequency stratum.
Evaluation reactions
were not pooled across template partitions as independent samples.
Reported mean gains were arithmetic means of the three within-partition
accuracy differences.
Template-subset definitions and reporting
conventions are provided in \noteref{1}.

\subsection{Multistep planning and
evaluation}\label{multistep-planning-and-evaluation}

Multistep evaluation used \name{Pistachio Hard}\citep{yu2024doubleended} (100
targets), \name{USPTO-190}\citep{chen2020retrostar} (190 targets) and \name{PatMol-150}
(150 targets).
\name{Pistachio Hard} was evaluated without its original
specified-starting-material constraint.
\name{PatMol-150} was constructed from
2025 public patents in \name{DianShi-RxnDB} and contained reference routes of
7--13 steps.
After molecular standardization, duplicate targets were
removed, and targets were excluded if they appeared among products in
\name{Train-Full}, targets in either public benchmark or molecules in the
starting-material inventory.
Benchmark details and the \name{PatMol-150}
construction procedure are provided in \noteref{7}.

We evaluated three planning frameworks: \name{RetroAgent}\citep{zhu2026retroagent},
\name{Retro-R1-modified}\citep{liu2025retror1} and
\name{DianShi-RetroPlanSystem-v0}.
All three used \name{Qwen3.5-397B} as the
planner model and a common RDKit-standardized inventory of 23,081,629
commercially available eMolecules building blocks\citep{emolecules}.
Within each framework, the planner
model, inventory, Top-10 single-step candidate setting and search protocol
were held fixed across single-step model comparisons.
\name{Retro-R1-modified} adapted the original inference protocol for a
general-purpose planner model.
\name{DianShi-RetroPlanSystem-v0} represented
search as an AND--OR tree, in which a molecule node closed when at least one
candidate reaction closed and a reaction node closed only when all
required precursor molecules were available or solved.
Protocol details, including the tool interface, closure propagation and cycle handling, are provided in \noteref{7}.

All evaluations applied a common \name{AOT*-style} template-based
reaction-validation procedure\citep{song2025aotstar} after single-step
prediction, with each model's corresponding template library and
rdchiral\citep{coley2019rdchiral} for template application
(\name{Train-Full} for the DianShi models; USPTO-derived libraries for \name{RSGPT}, \name{RXNGraphormer} and the default planning models).
Candidates rewritten to regenerate the target
product, or any of its components, were discarded.
Template-library sources, matching thresholds and fallback
settings are provided in \noteref{7}.

Performance was assessed as target-level solve rate, defined as the
fraction of targets for which at least one complete route, with all
terminal precursors in the starting-material inventory, was found
within the search budget.
Multiple routes
for one target counted as one success.
Budget \(N\) counted newly
expanded distinct molecules in \name{RetroAgent}, single-step tool calls in
\name{Retro-R1-modified}, and expanded molecule nodes in
\name{DianShi-RetroPlanSystem-v0}, where each expansion invoked the
single-step model once.
Results at \(N=50,100,200,300,400\) and \(500\)
were budget levels along the same search trajectory.
Single-step model
comparisons were performed within each framework.
Solve-rate
trajectories are provided in \noteref{7} and \edfigref{5}.

%% file: sections/05-data-availability.tex
\section{Data availability}\label{data-availability}

{\sloppy
The DianShi-curated datasets and derived resources are available at
\url{https://github.com/DianShiGroup/DianShi-RetroStepLM} and
\url{https://huggingface.co/datasets/DianShiGroup/DianShi-RetroPlanSystem-Assets}
under a Creative Commons Attribution 4.0 International License (CC BY 4.0;
\url{https://creativecommons.org/licenses/by/4.0/}).
They comprise the evaluation partitions
\name{DianShi-PatentRxnStep-Test-2001-2024} (3,798 reactions),
\name{DianShi-PatentRxnStep-Test-2025} (1,886 reactions),
\name{DianShi-PatentRxnStep-Val-2001-2024} (3,798 reactions) and
\name{DianShi-PatentRxnStep-Val-2025} (1,886 reactions), the \name{PatMol-150}
multistep benchmark (150 targets), the \name{Train-Full} template library (the
library used for reaction validation) and the template-coverage cutoff scheme
(the per-partition retained and withheld template assignments for seeds 42, 43
and 44, together with the generation procedure).\par
}

%% file: sections/06-code-availability.tex
\section{Code availability}\label{code-availability}

The \name{DianShi-RetroStepLM-Direct} checkpoint
(\url{https://huggingface.co/DianShiGroup/DianShi-RetroStepLM-Direct}) and the
\name{DianShi-RetroPlanSystem-v0} multistep search framework
(\url{https://github.com/DianShiGroup/DianShi-RetroStepLM}) are available
under the Apache License 2.0
(\url{https://www.apache.org/licenses/LICENSE-2.0}).

%% file: sections/07-acknowledgements.tex
\section{Acknowledgments}

This project was supported by Shanghai Artificial Intelligence Laboratory.

%% file: sections/11-extended-data.tex
\section{\texorpdfstring{Extended Data for \emph{Reaction precedents extend retrosynthetic prediction beyond training coverage}}{Extended Data}}\label{extended-data}

\edfighead{1}{Top-k accuracy under template restriction and retrieval corpus expansion}

\begin{figure}[H]
\centering
\includegraphics[width=\linewidth]{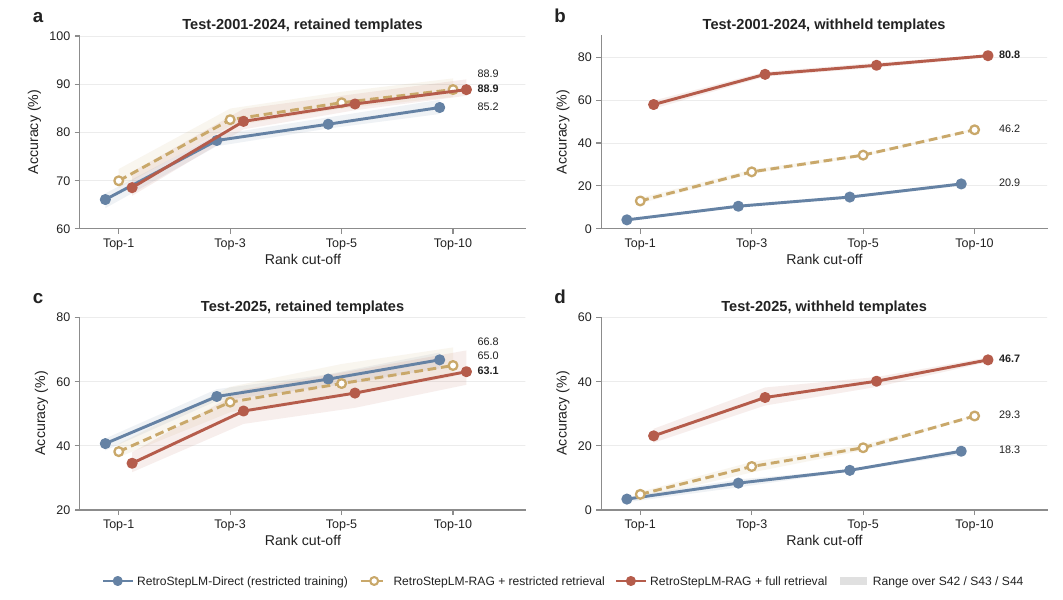}
\end{figure}

\textbf{a--d}, Complete precursor-set exact-match accuracy at the Top-1, Top-3, Top-5 and Top-10 rank cut-offs on \name{Test-2001-2024} (\textbf{a}, retained; \textbf{b}, withheld templates) and \name{Test-2025} (\textbf{c}, retained; \textbf{d}, withheld templates).
Points are arithmetic means over the three template partitions (S42, S43, S44), with light shading spanning the partitions and reactions not pooled across them.
Results are reported in \noteref{4}.

\edfighead{2}{Prediction accuracy across precedent-frequency ranges}

\begin{figure}[H]
\centering
\includegraphics[width=\linewidth]{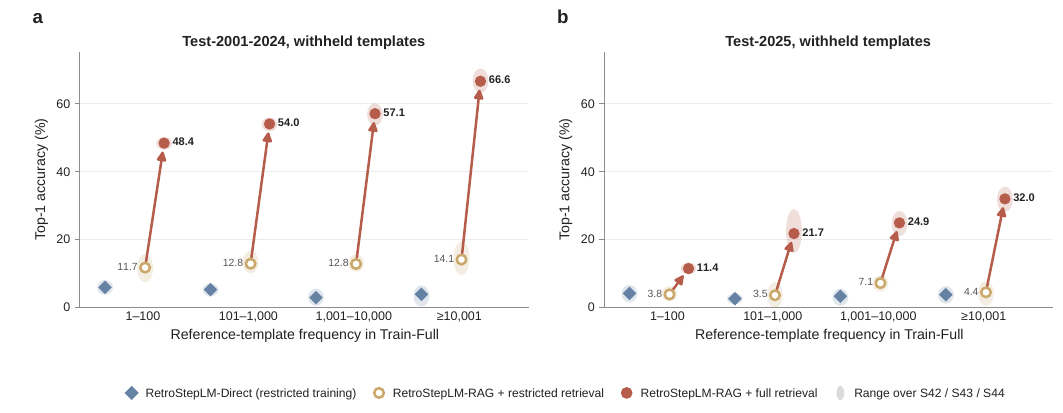}
\end{figure}

\textbf{a,b}, Top-1 accuracy on withheld-template reactions in \name{Test-2001-2024} (\textbf{a}) and \name{Test-2025} (\textbf{b}), stratified by the frequency of the recorded local reaction template in \name{Train-Full}: 1--100, 101--1,000, 1,001--10,000 and $\geq$10,001 records.
Points are arithmetic means over the three template partitions (S42, S43, S44), with light shading spanning the partitions and reactions not pooled across them.
Template counts, reaction denominators and results are reported in \noteref{4}.

\edfighead{3}{Top-k accuracy under temporal restriction and retrieval corpus expansion}

\begin{figure}[H]
\centering
\includegraphics[width=\linewidth]{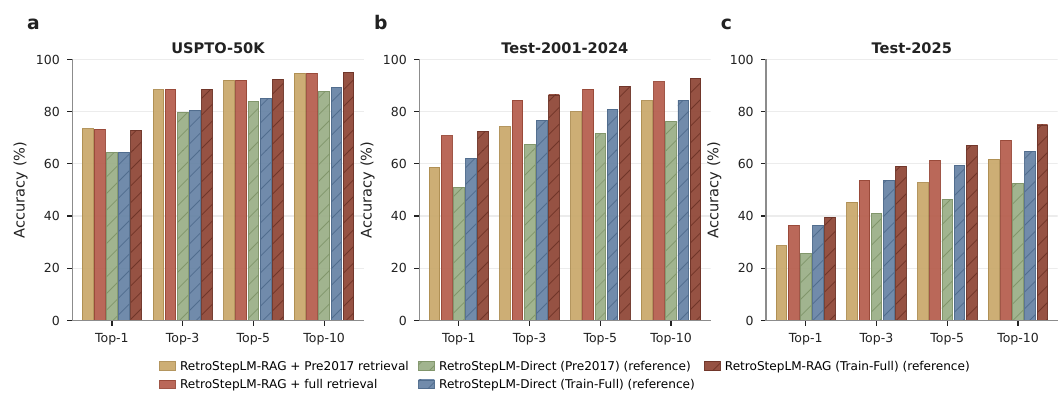}
\end{figure}

\textbf{a--c}, Complete precursor-set exact-match accuracy at the Top-1, Top-3, Top-5 and Top-10 rank cut-offs on \name{USPTO-50K} (\textbf{a}), \name{Test-2001-2024} (\textbf{b}) and \name{Test-2025} (\textbf{c}).
Complete results are reported in \noteref{4}.

\edfighead{4}{Prediction accuracy under different precedent-selection conditions}

\begin{figure}[H]
\centering
\includegraphics[width=\linewidth]{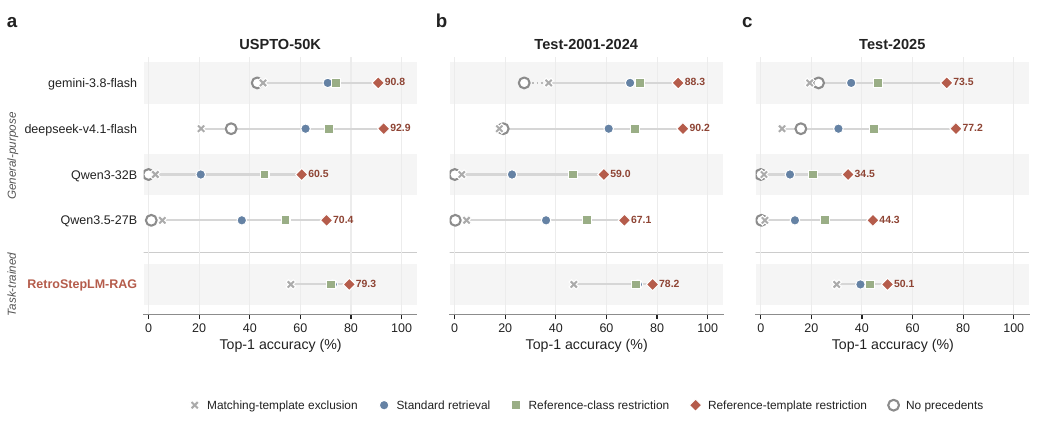}
\end{figure}

\textbf{a--c}, Top-1 accuracy on \name{USPTO-50K} (\textbf{a}), \name{Test-2001-2024} (\textbf{b}) and \name{Test-2025} (\textbf{c}) for four representative general-purpose models and \name{RetroStepLM-RAG} trained on \name{Train-Full}.
Conditions use the full DianShi retrieval corpus with standard retrieval, reference-class restriction, reference-template restriction or matching-template exclusion; grey circles mark the no-precedent condition for the general-purpose models.
Reference classes and templates derive from the recorded answer, making these privileged diagnostics.
The corpus and model remain fixed across selection conditions.
Protocols and complete results are reported in \noteref{5}.

\edfighead{5}{Multistep solve rates across search budgets}

\begin{figure}[H]
\centering
\includegraphics[width=\linewidth]{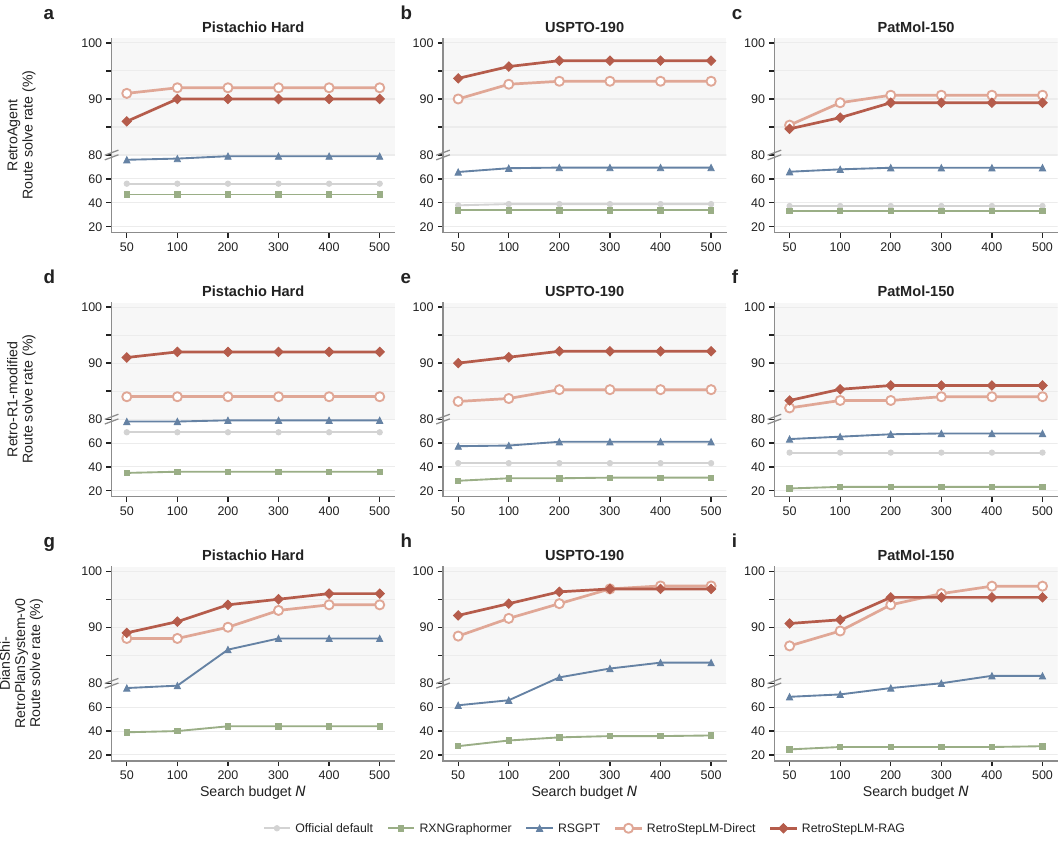}
\end{figure}

\textbf{a--i}, Route solve rates within \name{RetroAgent} (\textbf{a--c}), \name{Retro-R1-modified} (\textbf{d--f}) and \name{DianShi-RetroPlanSystem-v0} (\textbf{g--i}) on \name{Pistachio Hard} (100 targets), \name{USPTO-190} (190) and \name{PatMol-150} (150).
Within each framework, planner, starting-material inventory and search protocol are fixed.
Protocols and complete results are reported in \noteref{7}.

%% file: sections/12-supplementary-information.tex
\section{\texorpdfstring{Supplementary Information for \emph{Reaction precedents extend retrosynthetic prediction beyond training coverage}}{Supplementary Information}}\label{supplementary-information}

\sinotehead{1}{Dataset nomenclature and experimental configurations}

\sinotesub{1.1}{Dataset partitions}

\name{DianShi-PatentRxnStep-Dataset} comprises five base partitions, summarized in \tabref{1}.
Years refer to the year of the source patent recorded in \name{DianShi-RxnDB}\citep{wang2026dianshirxndb}.
The five partitions are mutually disjoint by standardized target product.
Their membership remains fixed across training and retrieval configurations.

\sitablehead{1}{Dataset partitions.}

{\footnotesize
\begin{longtable}{@{}l l c r p{4.2cm}@{}}
\toprule
Partition & Dataset identifier & Recorded years & Reactions & Role \\
\midrule
\endhead
\bottomrule
\endlastfoot
\name{Train-Full} & \texttt{\scriptsize DianShi-PatentRxnStep-Train-Full} & 1976--2024 & 2,712,259 & Full training partition; also supplies the full retrieval corpus \\
\name{Val-2001-2024} & \texttt{\scriptsize DianShi-PatentRxnStep-Val-2001-2024} & 2001--2024 & 3,798 & Validation on historical records \\
\name{Test-2001-2024} & \texttt{\scriptsize DianShi-PatentRxnStep-Test-2001-2024} & 2001--2024 & 3,798 & Evaluation on historical records \\
\name{Val-2025} & \texttt{\scriptsize DianShi-PatentRxnStep-Val-2025} & 2025 & 1,886 & Temporally held-out validation \\
\name{Test-2025} & \texttt{\scriptsize DianShi-PatentRxnStep-Test-2025} & 2025 & 1,886 & Temporally held-out evaluation \\
\end{longtable}
}

The short partition names are used throughout this Supplementary Information.
\emph{Full} denotes the final training partition after the study's quality filtering and product-level exclusions, rather than the original database or the complete construction pool.
Dataset construction, filtering counts and partitioning procedures are described in \noteref{2}.

\sinotesub{1.2}{Training and retrieval configurations}

We distinguish the \textbf{training corpus}, which supplies reactions for task-specific parameter updates; the \textbf{checkpoint}, which contains the resulting model parameters; the \textbf{retrieval corpus}, which contains reaction records available for precedent selection; the \textbf{retrieval index}, which implements search over that corpus; and the \textbf{retrieved precedents}, which are the selected examples supplied to the model for a particular query.
The \textbf{restricted retrieval corpus} matches a coverage-restricted training corpus, whereas the \textbf{full retrieval corpus} comprises all \name{Train-Full} reactions.
\emph{Restricted} and \emph{full} qualify corpus coverage only: the retrieval procedure is identical across coverage conditions, and procedure-level constraints on precedent selection are described in \noteref{5}.
A record set can serve both training and retrieval roles without becoming a separate dataset.

Three coverage labels identify the DianShi training and retrieval resources used in the coverage experiments:

{\sloppy
\begin{itemize}
\item \textbf{Full} refers to \name{Train-Full}.
\item \textbf{\name{Pre2017}} refers to the subset of \name{Train-Full} containing records before 2017, identified as \texttt{DianShi-PatentRxnStep-Train-Pre2017}.
\item \textbf{\name{TemplateRestricted-S42}}, \textbf{\name{TemplateRestricted-S43}} and \textbf{\name{TemplateRestricted-S44}} refer to subsets of \name{Train-Full} obtained by removing all reactions bearing the templates withheld in the corresponding template partition.
Their identifiers are \texttt{DianShi-PatentRxnStep-Train-\allowbreak TemplateRestricted-\allowbreak S42}, \texttt{DianShi-PatentRxnStep-Train-\allowbreak TemplateRestricted-\allowbreak S43} and \texttt{DianShi-PatentRxnStep-Train-\allowbreak TemplateRestricted-\allowbreak S44}.
\end{itemize}
}

For each restricted \name{RetroStepLM-RAG} configuration, the restricted record set supplied both the fine-tuning reactions and the matching restricted retrieval corpus.
After training, the same checkpoint was evaluated with either that restricted retrieval corpus or the full retrieval corpus.
\sinotesub{1.3}{Template-defined evaluation subsets and reporting}

The template intervention uses three alternative template partitions, denoted S42, S43 and S44.
Within each partition, approximately half of the distinct reaction templates represented in the two base test sets were designated as withheld.
Templates shared between the two test sets received the same assignment.

Each base test set was divided into \textbf{retained-template} and \textbf{withheld-template} subsets according to the corresponding partition.
Retained templates had at least one reaction in the restricted training corpus and its matching restricted retrieval corpus; withheld templates had none.
Temporal and transformation-based shifts represent distinct evaluation axes\citep{yu2024retroood}.

{\sloppy
Subset display names combine the evaluation window, template status and partition, for example, \emph{2001--2024 / Withheld (partition seed 42)}.
The corresponding dataset identifier is \texttt{DianShi-PatentRxnStep-Test-2001-2024-\allowbreak TemplateWithheld-S42}; retained subsets use \texttt{TemplateRetained}.\par
}

Within each partition, the retained- and withheld-template subsets are mutually exclusive and together comprise the corresponding base test set.

Single-step accuracy measures exact agreement with the complete standardized reference precursor set, following the scoring procedure in \secref{methods}.
Accuracy within a subset uses the number of evaluation reactions in that subset as its denominator.
Cross-partition summaries are arithmetic means of the three partition-specific accuracies, calculated separately for each test set, template status and, where applicable, precedent-frequency stratum.
Evaluation reactions are not pooled across partitions as independent samples.
A reported mean gain is the arithmetic mean of the three within-partition accuracy differences.

\sinotesub{1.4}{Configuration correspondence}

\tabref{2} summarizes the coverage experiments and their reference configurations.
Coverage labels refer to the record sets defined in \notesec{1.2}.
\name{RetroStepLM-Direct} receives no retrieved precedents.
For \name{RetroStepLM-RAG}, the retrieval corpus used during training matches its training corpus; the inference retrieval corpus is specified separately.

\sitablehead{2}{Configuration correspondence.}

{\footnotesize
\setlength{\tabcolsep}{3pt}
\begin{longtable}{@{}l l l p{3.7cm}@{}}
\toprule
Model & Training corpus & Inference retrieval corpus & Comparison role \\
\midrule
\endhead
\bottomrule
\endlastfoot
\name{RetroStepLM-Direct} & \texttt{\footnotesize TemplateRestricted-S\{s\}} & None & Separately trained template-restricted reference \\
\name{RetroStepLM-RAG} & \texttt{\footnotesize TemplateRestricted-S\{s\}} & \texttt{\footnotesize TemplateRestricted-S\{s\}} & Restricted-retrieval condition \\
Same \name{RetroStepLM-RAG} checkpoint & \texttt{\footnotesize TemplateRestricted-S\{s\}} & \texttt{\footnotesize Full} & Expanded-retrieval condition \\
\name{RetroStepLM-Direct} & \texttt{\footnotesize Pre2017} & None & Separately trained temporally restricted reference \\
\name{RetroStepLM-RAG} & \texttt{\footnotesize Pre2017} & \texttt{\footnotesize Pre2017} & Restricted-retrieval condition \\
Same \name{RetroStepLM-RAG} checkpoint & \texttt{\footnotesize Pre2017} & \texttt{\footnotesize Full} & Expanded-retrieval condition \\
\name{RetroStepLM-Direct} & \texttt{\footnotesize Full} & None & Separately trained full-corpus reference \\
\name{RetroStepLM-RAG} & \texttt{\footnotesize Full} & \texttt{\footnotesize Full} & Separately trained full-corpus reference \\
\end{longtable}
}

Here, $s \in \{42, 43, 44\}$; the three template-restricted rows apply separately to each partition.
Within each paired \name{RetroStepLM-RAG} comparison, the checkpoint, evaluation targets, retrieval procedure, precedent serialization, prompting, decoding and scoring remain fixed.
Only the retrieval corpus and its corresponding index change.
Standard inference supplies five retrieved precedents per query.

The full-corpus \name{RetroStepLM-RAG} reference has separately trained parameters.
It is therefore distinct from a restricted-training checkpoint evaluated with the full retrieval corpus.
General-purpose model configurations and additional adaptation and retrieval-transfer experiments are described in \cref{note:6,note:8}, respectively.

\sinotehead{2}{Reaction dataset construction and partitioning}

\sinotesub{2.1}{Reaction standardization and deduplication}

\name{DianShi-PatentRxnStep-Dataset} was derived from \name{DianShi-RxnDB}\citep{wang2026dianshirxndb}, a provenance-linked organic reaction-data platform that we previously developed.
\name{DianShi-RxnDB} uses a fully automated pipeline to extract structured reaction records from patent text, images and reaction schemes, primarily from USPTO and EPO organic-synthesis patents published between 1976 and 2025.
Each record represents a reported single-step experiment and links to its source patent.

Reactant and product SMILES were standardized using RDKit\citep{rdkit}.
Reaction-level duplicates were identified using a standardized reactant-set--product-set signature.
Atom-mapping numbers were removed from the comparison representation, and molecular components on each side were sorted deterministically, so differences in component order or mapping-number assignment did not produce distinct signatures.
Records with identical signatures were treated as duplicates.
When duplicate records spanned multiple years, the record with the earliest year recorded in \name{DianShi-RxnDB} was retained.

\sinotesub{2.2}{Reaction-quality filtering and template processing}

Reaction-quality filtering excluded records with multiple products, repeated molecules on either side, structural overlap between reactants and products, more than four reactant components or inorganic products.
Records were also excluded when atom mappings were inconsistent or reaction-centre or template extraction was unsuccessful.

Local reaction templates\citep{chen2021localretro} were grouped across the quality-controlled corpus.
Templates represented by fewer than six reactions were removed before temporal partitioning, following the cutoff used in previous template-frequency analyses\citep{tanovic2026bias}.
This threshold applied to template frequency in the corpus at that processing stage.

The principal processing and partitioning rules are summarized in \tabref{3}.

\sitablehead{3}{Processing and partitioning rules.}

\begin{longtable}{@{}p{3.8cm} p{7.1cm} p{3.8cm}@{}}
\toprule
Processing component & Rule & Scope \\
\midrule
\endhead
\bottomrule
\endlastfoot
Molecular standardization & Standardize reactant and product SMILES using RDKit & Reaction records entering dataset construction \\
Reaction deduplication & Compare standardized reactant-set--product-set signatures after removing mapping numbers and sorting components; retain the earliest recorded year for cross-year duplicates & Duplicate reaction records \\
Structural quality filtering & Exclude multiple products, repeated molecules, reactant--product overlap, more than four reactant components and inorganic products & Deduplicated records \\
Mapping and extraction filtering & Exclude inconsistent atom mappings and unsuccessful reaction-centre or template extraction & Records undergoing quality control \\
Template-frequency filtering & Remove local reaction templates represented by fewer than six reactions & Quality-controlled corpus, before temporal partitioning \\
External benchmark exclusions & Compare standardized target products against the specified external dataset partitions & Training and evaluation candidates, as detailed in \notesec{2.3} \\
Internal partition separation & Require standardized target products to be mutually disjoint across all five base partitions & Final training, validation and test partitions \\
\end{longtable}

\sinotesub{2.3}{Temporal partitioning and product-level exclusions}

For temporal evaluation\citep{bradshaw2025generalize,qian2023textreact}, partitions were defined using the year of the source patent recorded in \name{DianShi-RxnDB}.
Records before 2025 supplied the training and historical evaluation pools, whereas records from 2025 supplied the temporally held-out evaluation pool.

\name{Val-2001-2024} and \name{Test-2001-2024} were sampled from 2001--2024 records, with stratification by year and template frequency in the full training corpus.
\name{Val-2025} and \name{Test-2025} were sampled from the 2025 pool, with stratification by patent type and template frequency.
The five base partitions were constructed jointly and were mutually disjoint by standardized target product\citep{tanovic2026bias}.
Consequently, overlapping year ranges between training and historical evaluation did not entail shared target products.

Product-level exclusions controlled overlap with external comparison datasets.
Training candidates were screened against products in the validation and test partitions of \name{USPTO-50K}\citep{coley2017retrosim} and \name{USPTO-Full}\citep{dai2019gln}, and \name{USPTO-MIT}\citep{jin2017wln}.
DianShi validation and test candidates were additionally screened against products in the training partitions of these USPTO datasets and of the corpus used to pretrain \name{RXNGraphormer}\citep{xu2025rxngraphormer}.
These exclusions operated on standardized target products irrespective of the associated precursor set.

\sinotesub{2.4}{Final dataset composition}

The resulting dataset comprised 2,712,259 reactions in \name{Train-Full}, 3,798 reactions each in \name{Val-2001-2024} and \name{Test-2001-2024}, and 1,886 reactions each in \name{Val-2025} and \name{Test-2025}.
Dataset identifiers and partition roles are summarized in \noteref{1}.

\name{Train-Full} contains records spanning 1976--2024 and supplies all reaction records in the full DianShi retrieval corpus, without additional corpus-level subsampling.
The restricted training and retrieval corpora used in the coverage experiments were derived from this fixed partition, as described in \noteref{4}.

\sinotehead{3}{Precedent retrieval, intermediate-label supervision and model training}

\sinotesub{3.1}{Retrieval implementation and precedent selection}

Retrieval used the target product alone, without reference reaction-class or template information.
Products in each retrieval corpus were represented by extended-connectivity fingerprints (ECFPs)\citep{rogers2010ecfp}.
A binary flat index returned 200 candidate reactions ranked by Hamming distance to the query fingerprint.
These candidates were then reranked by the Tanimoto similarity between their product fingerprints and the query fingerprint.

During \name{RetroStepLM-RAG} fine-tuning, the query reaction, duplicate reactions and potentially answer-disclosing records from the same patent were excluded from candidate precedents.
At each training epoch, five eligible precedents were randomly selected from the ten highest-similarity candidates.
At standard inference, \name{RetroStepLM-RAG} received the five highest-ranked eligible precedents deterministically.
The general-purpose model evaluations used the same standard inference selection procedure.

Indices were constructed from the full training partition and each restricted training corpus without additional corpus-level subsampling.
Within each retrieval corpus expansion comparison, the checkpoint and retrieval procedure remained unchanged; inference switched between indices built from the corresponding restricted and full record sets.
Configuration correspondence is provided in \noteref{1}, and the interventions are detailed in \noteref{4}.
Reference-informed modifications to precedent selection are described separately in \noteref{5}.

\sinotesub{3.2}{Precedent representation and model inputs}

Each retrieved precedent contained standardized product and reactant SMILES, together with synthon SMILES and corresponding leaving groups derived from the atom-mapped reaction.
These fields represented the precedent reaction and were supplied as reference context.

\name{RetroStepLM-Direct} received only the query product.
\name{RetroStepLM-RAG} received the query product together with five retrieved precedents.
The query reaction's reference reactants, synthons and leaving groups were used to construct training targets and were not included in the query input.
At inference, these fields were generated by the model.

In the prompt records below, \texttt{\{products\}} denotes the query product and \texttt{\{rag\_examples\}} denotes the retrieved-precedent block.
The placeholders \texttt{\{synthons\_smiles\}}, \texttt{\{leaving\_groups\}} and \texttt{\{reactants\}} occur in the supervised target and denote labels derived from the query reaction.

\sinotesub{3.3}{Intermediate-label supervision and prompt templates}

Both DianShi models used intermediate-label supervision.
Synthon- and leaving-group-based decompositions have also been used in structured retrosynthesis models\citep{xiong2025reactseq,zhao2025multitask}.
Atom-mapped reactions supplied labels comprising synthons associated with the retrosynthetic disconnection, corresponding leaving groups and the final standardized reactant set.
The target sequence presented these fields in the order \texttt{synthons\_smiles}, \texttt{leaving\_groups} and \texttt{reactants}, with the final reactant string additionally enclosed in \texttt{\textless{}answer\textgreater{}} tags.

The structured prompt text and supervised target are shown separately below.
The \texttt{\textless{}think\textgreater{}} tags delimit the serialized intermediate fields.

\noindent\textbf{\name{RetroStepLM-Direct} input}

\begin{lstlisting}
---SYSTEM---
You are an expert chemist specializing in retrosynthetic analysis. Given the SMILES representation of a product molecule, your task is to predict the SMILES of the reactants required to synthesize it.
Input:
The product SMILES.
Task requirements (step-by-step reasoning is required):
1. Infer the synthons based on the structure of the product.
2. Identify the corresponding leaving groups associated with these synthons.
3. Reconstruct the most plausible reactant molecules by combining the synthons with appropriate leaving groups.
You must explicitly show your reasoning step by step for each of the above stages.
Please output the reasoning process, and write the final reactant SMILES strictly in the following format:
<answer>REACTANT_SMILES</answer>
If multiple reactants are required, separate their SMILES using a period (.).
---USER---
The product SMILES is {products}.
\end{lstlisting}

\noindent\textbf{\name{RetroStepLM-Direct} supervised target}

\begin{lstlisting}
<think>The product SMILES is {products}. By analyzing the product SMILES, the synthons involved in this reaction are identified as {synthons_smiles}.
Based on the product structure and the inferred synthons, the corresponding leaving groups for this reaction are deduced to be {leaving_groups}.
By combining the synthons with the identified leaving groups, the final reactant(s) are inferred as: {reactants}</think>
<answer>{reactants}</answer>
\end{lstlisting}

\noindent\textbf{\name{RetroStepLM-RAG} input text}

\begin{lstlisting}
You are an expert chemist specializing in retrosynthetic analysis. Given the SMILES representation of a product molecule, your task is to predict the SMILES of the reactants required to synthesize it. The input consists of a product SMILES and multiple referenced reactions. Task requirements (step-by-step reasoning is required):
1. Infer the synthons based on the structure of the product.
2. Identify the corresponding leaving groups associated with these synthons.
3. Reconstruct the most plausible reactant molecules by combining the synthons with appropriate leaving groups.
You must explicitly show your reasoning step by step for each of the above stages.
Please output the reasoning process, and write the final reactant SMILES strictly in the following format:
<answer>REACTANT_SMILES</answer>
If multiple reactants are required, separate their SMILES using a period (.).

The product SMILES is {products}. Below are similar reactions retrieved from a reaction database for reference:
{rag_examples}
\end{lstlisting}

\noindent\textbf{\name{RetroStepLM-RAG} supervised target}

\begin{lstlisting}
<think>The product SMILES is {products}. By analyzing the product SMILES and multiple referenced reactions, the synthons involved in this reaction are identified as {synthons_smiles}.
Based on the product structure and the inferred synthons, the corresponding leaving groups for this reaction are deduced to be {leaving_groups}.
By combining the synthons with the identified leaving groups, the final reactant(s) are inferred as: {reactants}</think>
<answer>{reactants}</answer>
\end{lstlisting}

Training minimized the autoregressive negative log-likelihood over the complete supervised target sequence, as defined in \secref{methods}.
Only the final reactant set contributed to precursor-set accuracy.
General-purpose model prompting and model-specific call settings are described in \noteref{6}.

\sinotesub{3.4}{Model training}

\name{DianShi-RetroStepLM-Direct} and \name{DianShi-RetroStepLM-RAG} were developed from \name{Qwen3-4B-Base}\citep{yang2025qwen3} using full-parameter supervised fine-tuning; for \name{RetroStepLM-RAG}, this took the form of retrieval-conditioned supervised fine-tuning, with the target product and five retrieved precedents supplied as input.
SMILES were space-segmented into atom- and bond-level tokens following the Molecular Transformer tokenization scheme\citep{schwaller2019moleculartransformer} and encoded using the native Qwen tokenizer without vocabulary expansion.

Full-corpus models used \name{Train-Full}.
Coverage-intervention models used their corresponding \name{Pre2017} or \name{TemplateRestricted-S\{s\}} training corpus, where $s \in \{42, 43, 44\}$.
For each \name{RetroStepLM-RAG} model, training precedents were retrieved from the same record set that supplied its fine-tuning reactions.

Training used 32 H200 GPUs across four nodes, with bfloat16 precision, FlashAttention, DeepSpeed ZeRO-3 and gradient checkpointing.
The recorded training settings are summarized in \tabref{4}.

\sitablehead{4}{Training settings.}

\begin{longtable}{@{}p{5.6cm} p{9.3cm}@{}}
\toprule
Item & Setting \\
\midrule
\endhead
\bottomrule
\endlastfoot
Base model & \name{Qwen3-4B-Base} \\
Parameter updates & Full-parameter supervised fine-tuning \\
Supervision & Synthon--leaving-group intermediate labels \\
Hardware & 32 H200 GPUs across four nodes \\
Precision & bfloat16 \\
Training implementation & FlashAttention; DeepSpeed ZeRO-3; gradient checkpointing \\
Per-device batch size & 4 \\
Gradient accumulation & 1 step \\
Effective batch size & 128 \\
Training duration & 10 epochs \\
Learning rate & $1 \times 10^{-4}$ \\
Learning-rate schedule & Cosine decay with linear warm-up over the first 5\% of training steps \\
Tokenization & Molecular Transformer-style SMILES spacing followed by the native Qwen tokenizer; no vocabulary expansion \\
\name{RetroStepLM-RAG} training precedents & Random five from the ten highest-similarity eligible candidates per epoch \\
\name{RetroStepLM-RAG} standard inference precedents & Deterministic Top-5 \\
\end{longtable}

Additional adaptation experiments and training-time precedent-sampling ablations are described in \noteref{8}.
Complete single-step performance comparisons are reported in \noteref{6}.

\sinotesub{3.5}{Decoding and candidate extraction}

At inference, both DianShi models used beam search to generate ranked precursor-set candidates for Top-1, Top-3, Top-5 and Top-10 evaluation.
The final reactant string enclosed in \texttt{\textless{}answer\textgreater{}} tags was extracted from each generated sequence.

Predicted and reference precursor sets were standardized using RDKit, converted to canonical SMILES and compared after deterministic component sorting\citep{maziarz2025syntheseus}.
Component order therefore did not affect correctness.
Candidates were evaluated in their generated rank order.
Missing predictions, invalid molecular strings and partial precursor-set matches were counted as incorrect under the common scoring procedure defined in \secref{methods}.

\sinotehead{4}{Fixed-parameter retrieval corpus expansion}

\sinotesub{4.1}{Template-coverage intervention}

We constructed three alternative template partitions using partition seeds 42, 43 and 44.
Within each partition, approximately half of the distinct local reaction templates represented in \name{Test-2001-2024} and \name{Test-2025} were designated as withheld.
Templates shared between the two test sets received the same retain-or-withhold assignment.
The withholding target referred to distinct templates rather than reaction counts.

All reactions bearing withheld templates were removed from \name{Train-Full}.
The remaining records served as both the \name{TemplateRestricted-S\{s\}} training corpus and its matching restricted retrieval corpus, where $s \in \{42, 43, 44\}$.
Withheld evaluation templates therefore had zero occurrence in both restricted resources, whereas retained evaluation templates had at least one corresponding training reaction.
Restricted-corpus sizes are summarized in \tabref{5}.

\sitablehead{5}{Template-restricted training and retrieval corpora.}

\begin{longtable}{@{}l r r r@{}}
\toprule
Record set & Reactions retained & Percentage of \name{Train-Full} (\%) & Reactions removed \\
\midrule
\endhead
\bottomrule
\endlastfoot
\name{Train-Full} & 2,712,259 & 100.00 & 0 \\
\name{TemplateRestricted-S42} & 1,804,926 & 66.55 & 907,333 \\
\name{TemplateRestricted-S43} & 1,673,416 & 61.70 & 1,038,843 \\
\name{TemplateRestricted-S44} & 1,564,454 & 57.68 & 1,147,805 \\
\end{longtable}

For each partition, \name{RetroStepLM-Direct} and \name{RetroStepLM-RAG} were trained on the restricted corpus.
\name{RetroStepLM-RAG} training used precedents retrieved from the matching restricted retrieval corpus.
After training, its checkpoint was fixed and evaluated with either that restricted retrieval corpus or the full retrieval corpus.
The two conditions used identical evaluation targets and execution settings apart from the retrieval corpus and corresponding index.
\name{RetroStepLM-Direct} provided a separately trained reference.

Each base test set was divided into retained- and withheld-template subsets according to the corresponding partition.
These assignments remained unchanged after retrieval corpus expansion.
Subset sizes are reported in \tabref{6}.

\sitablehead{6}{Template-defined evaluation subset sizes.}
Entries are numbers of evaluation reactions.

\begin{longtable}{@{}l r r r r@{}}
\toprule
Partition & 2001--2024 / Retained & 2001--2024 / Withheld & 2025 / Retained & 2025 / Withheld \\
\midrule
\endhead
\bottomrule
\endlastfoot
S42 & 1,977 & 1,821 & 970 & 916 \\
S43 & 1,842 & 1,956 & 937 & 949 \\
S44 & 1,797 & 2,001 & 878 & 1,008 \\
\end{longtable}

Within each partition, the retained and withheld subsets contained 671 distinct templates each for \name{Test-2001-2024}, and 426 and 427 distinct templates, respectively, for \name{Test-2025}.

\tabref{7} reports complete precursor-set exact-match accuracy.
All models in a given partition were trained on its \name{TemplateRestricted-S\{s\}} corpus.
\emph{Restricted retrieval} and \emph{full retrieval} denote evaluations of the same \name{RetroStepLM-RAG} checkpoint.

\sitablehead{7}{Complete template-intervention results (\%).}

\noindent\textbf{a, \name{Test-2001-2024} / Retained.}

{\small
\begin{longtable}{@{}l l r r r r@{}}
\toprule
Partition & Configuration & Top-1 & Top-3 & Top-5 & Top-10 \\
\midrule
\endhead
\bottomrule
\endlastfoot
S42 & \name{RetroStepLM-Direct} & 67.32 & 79.46 & 83.11 & 86.75 \\
S42 & \name{RetroStepLM-RAG}, restricted retrieval & 72.38 & 84.93 & 88.42 & 91.20 \\
S42 & \name{RetroStepLM-RAG}, full retrieval & 71.78 & 84.88 & 87.96 & 91.00 \\
S43 & \name{RetroStepLM-Direct} & 66.56 & 78.34 & 81.16 & 84.91 \\
S43 & \name{RetroStepLM-RAG}, restricted retrieval & 69.00 & 81.65 & 84.96 & 87.35 \\
S43 & \name{RetroStepLM-RAG}, full retrieval & 66.94 & 81.32 & 85.12 & 88.00 \\
S44 & \name{RetroStepLM-Direct} & 64.27 & 77.13 & 80.80 & 83.81 \\
S44 & \name{RetroStepLM-RAG}, restricted retrieval & 68.45 & 81.30 & 85.09 & 88.09 \\
S44 & \name{RetroStepLM-RAG}, full retrieval & 66.83 & 80.63 & 84.59 & 87.59 \\
\end{longtable}
}

\noindent\textbf{b, \name{Test-2001-2024} / Withheld.}

{\small
\begin{longtable}{@{}l l r r r r@{}}
\toprule
Partition & Configuration & Top-1 & Top-3 & Top-5 & Top-10 \\
\midrule
\endhead
\bottomrule
\endlastfoot
S42 & \name{RetroStepLM-Direct} & 4.39 & 11.26 & 15.54 & 21.25 \\
S42 & \name{RetroStepLM-RAG}, restricted retrieval & 11.81 & 25.65 & 33.66 & 46.07 \\
S42 & \name{RetroStepLM-RAG}, full retrieval & 56.07 & 71.00 & 75.23 & 79.96 \\
S43 & \name{RetroStepLM-Direct} & 4.24 & 10.12 & 14.57 & 21.47 \\
S43 & \name{RetroStepLM-RAG}, restricted retrieval & 14.37 & 27.66 & 34.92 & 45.76 \\
S43 & \name{RetroStepLM-RAG}, full retrieval & 58.08 & 72.03 & 76.23 & 81.19 \\
S44 & \name{RetroStepLM-Direct} & 3.75 & 10.09 & 14.19 & 19.99 \\
S44 & \name{RetroStepLM-RAG}, restricted retrieval & 12.69 & 26.34 & 34.48 & 46.73 \\
S44 & \name{RetroStepLM-RAG}, full retrieval & 59.82 & 73.26 & 77.51 & 81.26 \\
\end{longtable}
}

\noindent\textbf{c, \name{Test-2025} / Retained.}

{\small
\begin{longtable}{@{}l l r r r r@{}}
\toprule
Partition & Configuration & Top-1 & Top-3 & Top-5 & Top-10 \\
\midrule
\endhead
\bottomrule
\endlastfoot
S42 & \name{RetroStepLM-Direct} & 41.24 & 57.63 & 62.27 & 67.11 \\
S42 & \name{RetroStepLM-RAG}, restricted retrieval & 40.62 & 58.35 & 65.46 & 70.62 \\
S42 & \name{RetroStepLM-RAG}, full retrieval & 37.94 & 57.53 & 63.51 & 69.69 \\
S43 & \name{RetroStepLM-Direct} & 38.53 & 54.43 & 61.58 & 69.05 \\
S43 & \name{RetroStepLM-RAG}, restricted retrieval & 37.46 & 51.97 & 57.10 & 62.86 \\
S43 & \name{RetroStepLM-RAG}, full retrieval & 33.94 & 48.13 & 53.90 & 60.51 \\
S44 & \name{RetroStepLM-Direct} & 42.26 & 53.99 & 58.43 & 64.12 \\
S44 & \name{RetroStepLM-RAG}, restricted retrieval & 36.45 & 50.46 & 55.58 & 61.50 \\
S44 & \name{RetroStepLM-RAG}, full retrieval & 31.78 & 46.81 & 51.82 & 59.00 \\
\end{longtable}
}

\noindent\textbf{d, \name{Test-2025} / Withheld.}

{\small
\begin{longtable}{@{}l l r r r r@{}}
\toprule
Partition & Configuration & Top-1 & Top-3 & Top-5 & Top-10 \\
\midrule
\endhead
\bottomrule
\endlastfoot
S42 & \name{RetroStepLM-Direct} & 3.71 & 8.52 & 12.55 & 18.78 \\
S42 & \name{RetroStepLM-RAG}, restricted retrieval & 3.71 & 11.68 & 18.56 & 29.15 \\
S42 & \name{RetroStepLM-RAG}, full retrieval & 20.96 & 32.53 & 38.32 & 46.07 \\
S43 & \name{RetroStepLM-Direct} & 2.42 & 7.06 & 12.01 & 18.86 \\
S43 & \name{RetroStepLM-RAG}, restricted retrieval & 5.27 & 14.96 & 20.34 & 29.50 \\
S43 & \name{RetroStepLM-RAG}, full retrieval & 25.29 & 38.15 & 41.41 & 47.52 \\
S44 & \name{RetroStepLM-Direct} & 4.07 & 9.52 & 12.50 & 17.26 \\
S44 & \name{RetroStepLM-RAG}, restricted retrieval & 5.65 & 13.99 & 19.25 & 29.17 \\
S44 & \name{RetroStepLM-RAG}, full retrieval & 23.02 & 34.42 & 40.58 & 46.63 \\
\end{longtable}
}

Retrieval corpus expansion improved withheld-template Top-1 accuracy in every partition.
Mean partition-level accuracy increased from 13.0\% to 58.0\% on \name{Test-2001-2024} and from 4.9\% to 23.1\% on \name{Test-2025}.
Retained-template accuracy decreased in every partition, with corresponding means changing from 69.9\% to 68.5\% and from 38.2\% to 34.6\%.
\sinotesub{4.2}{Precedent-frequency stratification}

We stratified withheld-template evaluation reactions by the number of records bearing their local reaction template in \name{Train-Full}, which also constituted the full retrieval corpus.
Frequencies were calculated before template withholding.
The four frequency ranges were 1--100, 101--1,000, 1,001--10,000 and at least 10,001 reactions per template.
\tabref{8} reports distinct-template counts, evaluation-reaction counts and Top-1 accuracy for each partition and stratum.
Evaluation-reaction counts provide the accuracy denominators.
All three columns within a row used models trained on the corresponding \name{TemplateRestricted-S\{s\}} corpus; the two \name{RetroStepLM-RAG} columns share a checkpoint.

\sitablehead{8}{Withheld-template composition and frequency-stratified Top-1 accuracy (\%).}

\noindent\textbf{a, \name{Test-2001-2024} / Withheld.}

{\footnotesize
\setlength{\tabcolsep}{3pt}
\begin{longtable}{@{}p{2.2cm} l >{\raggedleft\arraybackslash}p{1.55cm} >{\raggedleft\arraybackslash}p{1.55cm} >{\raggedleft\arraybackslash}p{1.5cm} >{\raggedleft\arraybackslash}p{2.35cm} >{\raggedleft\arraybackslash}p{2.35cm}@{}}
\toprule
Reactions per template in full corpus & Partition & Distinct templates & Evaluation reactions & \name{RetroStepLM-Direct} & \name{RetroStepLM-RAG}, restricted retrieval & \name{RetroStepLM-RAG}, full retrieval \\
\midrule
\endhead
\bottomrule
\endlastfoot
1--100 & S42 & 290 & 305 & 6.56 & 8.85 & 48.52 \\
1--100 & S43 & 290 & 309 & 5.18 & 11.97 & 47.90 \\
1--100 & S44 & 290 & 308 & 5.84 & 14.29 & 48.70 \\
101--1,000 & S42 & 269 & 446 & 4.71 & 11.88 & 53.36 \\
101--1,000 & S43 & 269 & 452 & 5.53 & 11.50 & 54.42 \\
101--1,000 & S44 & 269 & 429 & 5.36 & 15.15 & 54.31 \\
1,001--10,000 & S42 & 92 & 592 & 2.70 & 11.99 & 55.07 \\
1,001--10,000 & S43 & 92 & 574 & 1.92 & 14.11 & 57.49 \\
1,001--10,000 & S44 & 92 & 551 & 3.99 & 12.16 & 58.62 \\
$\geq$10,001 & S42 & 20 & 478 & 4.81 & 13.39 & 64.64 \\
$\geq$10,001 & S43 & 20 & 621 & 4.99 & 17.87 & 66.34 \\
$\geq$10,001 & S44 & 20 & 713 & 1.68 & 10.94 & 68.86 \\
\end{longtable}
}

\noindent\textbf{b, \name{Test-2025} / Withheld.}

{\footnotesize
\setlength{\tabcolsep}{3pt}
\begin{longtable}{@{}p{2.2cm} l >{\raggedleft\arraybackslash}p{1.55cm} >{\raggedleft\arraybackslash}p{1.55cm} >{\raggedleft\arraybackslash}p{1.5cm} >{\raggedleft\arraybackslash}p{2.35cm} >{\raggedleft\arraybackslash}p{2.35cm}@{}}
\toprule
Reactions per template in full corpus & Partition & Distinct templates & Evaluation reactions & \name{RetroStepLM-Direct} & \name{RetroStepLM-RAG}, restricted retrieval & \name{RetroStepLM-RAG}, full retrieval \\
\midrule
\endhead
\bottomrule
\endlastfoot
1--100 & S42 & 188 & 212 & 4.25 & 4.72 & 11.32 \\
1--100 & S43 & 188 & 204 & 2.94 & 3.43 & 11.27 \\
1--100 & S44 & 188 & 214 & 5.14 & 3.27 & 11.68 \\
101--1,000 & S42 & 147 & 222 & 2.70 & 1.35 & 20.27 \\
101--1,000 & S43 & 147 & 211 & 2.37 & 3.32 & 27.49 \\
101--1,000 & S44 & 147 & 235 & 2.55 & 5.96 & 17.45 \\
1,001--10,000 & S42 & 72 & 276 & 3.99 & 6.16 & 22.46 \\
1,001--10,000 & S43 & 72 & 261 & 1.92 & 7.28 & 25.29 \\
1,001--10,000 & S44 & 72 & 264 & 3.79 & 7.95 & 26.89 \\
$\geq$10,001 & S42 & 20 & 206 & 3.88 & 1.94 & 29.61 \\
$\geq$10,001 & S43 & 20 & 273 & 2.56 & 6.23 & 34.07 \\
$\geq$10,001 & S44 & 20 & 295 & 4.75 & 5.08 & 32.20 \\
\end{longtable}
}

Expansion improved Top-1 accuracy in every partition and frequency range in both test sets.
For templates represented by 1--100 reactions in the full corpus, the mean partition-level gain was 36.7 percentage points on \name{Test-2001-2024} and 7.6 percentage points on \name{Test-2025}.

\sinotesub{4.3}{Temporal-coverage intervention}

The \name{Pre2017} training corpus retained 1,720,962 reactions before 2017, representing 63.45\% of \name{Train-Full}.
The remaining 991,297 reactions were excluded from both task-specific training and the \name{Pre2017} retrieval corpus.

\name{RetroStepLM-RAG} was trained using \name{Pre2017} reactions and precedents retrieved from the matching corpus.
After training, its checkpoint was fixed and evaluated with either the \name{Pre2017} retrieval corpus or the full retrieval corpus.

\tabref{9} reports complete precursor-set exact-match accuracy.
The two \name{Pre2017-trained} \name{RetroStepLM-RAG} rows share the same checkpoint.
\name{Pre2017-trained} \name{RetroStepLM-Direct} and the full-corpus models are separately trained references.

\sitablehead{9}{Complete temporal-intervention results (\%).}

\noindent\textbf{a, \name{USPTO-50K}.}

{\small
\begin{longtable}{@{}l l l r r r r@{}}
\toprule
Model & Training corpus & Inference retrieval corpus & Top-1 & Top-3 & Top-5 & Top-10 \\
\midrule
\endhead
\bottomrule
\endlastfoot
\name{RetroStepLM-Direct} & \name{Pre2017} & None & 64.2 & 79.8 & 84.0 & 87.7 \\
\name{RetroStepLM-Direct} & Full & None & 64.3 & 80.3 & 85.0 & 89.2 \\
\name{RetroStepLM-RAG} & \name{Pre2017} & \name{Pre2017} & 73.5 & 88.4 & 92.1 & 94.8 \\
\name{RetroStepLM-RAG} & \name{Pre2017} & Full & 73.3 & 88.5 & 92.0 & 94.7 \\
\name{RetroStepLM-RAG} & Full & Full & 72.8 & 88.6 & 92.2 & 94.9 \\
\end{longtable}
}

\noindent\textbf{b, \name{Test-2001-2024}.}

{\small
\begin{longtable}{@{}l l l r r r r@{}}
\toprule
Model & Training corpus & Inference retrieval corpus & Top-1 & Top-3 & Top-5 & Top-10 \\
\midrule
\endhead
\bottomrule
\endlastfoot
\name{RetroStepLM-Direct} & \name{Pre2017} & None & 50.8 & 67.3 & 71.6 & 76.3 \\
\name{RetroStepLM-Direct} & Full & None & 62.0 & 76.7 & 80.9 & 84.3 \\
\name{RetroStepLM-RAG} & \name{Pre2017} & \name{Pre2017} & 58.6 & 74.3 & 79.9 & 84.2 \\
\name{RetroStepLM-RAG} & \name{Pre2017} & Full & 70.7 & 84.4 & 88.6 & 91.5 \\
\name{RetroStepLM-RAG} & Full & Full & 72.4 & 86.4 & 89.8 & 92.7 \\
\end{longtable}
}

\noindent\textbf{c, \name{Test-2025}.}

{\small
\begin{longtable}{@{}l l l r r r r@{}}
\toprule
Model & Training corpus & Inference retrieval corpus & Top-1 & Top-3 & Top-5 & Top-10 \\
\midrule
\endhead
\bottomrule
\endlastfoot
\name{RetroStepLM-Direct} & \name{Pre2017} & None & 25.8 & 41.0 & 46.5 & 52.5 \\
\name{RetroStepLM-Direct} & Full & None & 36.4 & 53.6 & 59.4 & 64.7 \\
\name{RetroStepLM-RAG} & \name{Pre2017} & \name{Pre2017} & 28.7 & 45.4 & 52.7 & 61.8 \\
\name{RetroStepLM-RAG} & \name{Pre2017} & Full & 36.5 & 53.8 & 61.4 & 69.0 \\
\name{RetroStepLM-RAG} & Full & Full & 39.4 & 59.1 & 67.0 & 74.9 \\
\end{longtable}
}

Retrieval corpus expansion increased Top-1 accuracy by 12.1 percentage points on \name{Test-2001-2024} and 7.8 percentage points on \name{Test-2025}.
\name{USPTO-50K} accuracy remained essentially unchanged, from 73.5\% to 73.3\%.

With full retrieval, the \name{Pre2017-trained} \name{RetroStepLM-RAG} checkpoint exceeded full-corpus \name{RetroStepLM-Direct} on \name{Test-2001-2024} and achieved a similar Top-1 accuracy on \name{Test-2025}.
It remained below the separately trained full-corpus \name{RetroStepLM-RAG} reference on both test sets.

Expansion changed corpus size, recency and composition simultaneously; these comparisons do not separate their contributions.

\sinotehead{5}{Precedent-relevance diagnostics}

\sinotesub{5.1}{Models and evaluation settings}

We conducted reference-informed diagnostics to examine how precedent selection affected retrosynthetic prediction when the model and retrieval corpus were fixed.
Evaluations used \name{DianShi-RetroStepLM-RAG} trained on \name{Train-Full} and four representative general-purpose models: \name{gemini-3.8-flash}, \name{deepseek-v4.1-flash}, \name{Qwen3-32B} and \name{Qwen3.5-27B}.
All five models were evaluated on \name{USPTO-50K}, \name{Test-2001-2024} and \name{Test-2025}.

Each retrieval condition used the full DianShi retrieval corpus and supplied five precedents per query.
Within each model, parameters, evaluation targets, prompt structure, decoding and scoring settings remained fixed; only the precedent-selection rule changed.
\sinotesub{5.2}{Precedent-selection conditions}

Standard retrieval was compared with three reference-informed selection conditions.
Reaction classes and local reaction templates were obtained from the recorded reference reactions (\tabref{10}).

\sitablehead{10}{Precedent-selection rules used in the diagnostics.}

\begin{longtable}{@{}p{4.8cm} p{9.9cm}@{}}
\toprule
Condition & Selection rule \\
\midrule
\endhead
\bottomrule
\endlastfoot
Standard retrieval & Select precedents by target-product structural similarity, without reference reaction-class or template constraints \\
Reference-class restriction & Select precedents sharing the recorded reference reaction class \\
Reference-template restriction & Select precedents sharing the local reaction template of the reference reaction \\
Matching-template exclusion & Exclude precedents bearing the local reaction template of the reference reaction \\
\end{longtable}

Standard retrieval followed the procedure described in \noteref{3}.
In the matching-template exclusion condition, retrieval continued down the ranked candidate list to collect five eligible precedents.
When fewer than five remained, eligible precedents were cycled to fill the five context positions.
Thus, maintaining five positions did not necessarily imply five distinct precedents.

The reference-informed conditions used information derived from the query reaction's recorded answer.
They were therefore privileged diagnostics rather than retrieval settings assumed available for prospective prediction.

\sinotesub{5.3}{Diagnostic results}

\tabref{11} reports Top-1 complete precursor-set exact-match accuracy and corresponds to \edfigref{4}.
Conditions are defined in \tabref{10}.
A dash indicates that no corresponding result is reported.

\sitablehead{11}{Prediction accuracy under different precedent-selection conditions (\%).}

\noindent\textbf{a, \name{USPTO-50K}.}

{\footnotesize
\setlength{\tabcolsep}{3pt}
\begin{longtable}{@{}l >{\raggedleft\arraybackslash}p{2.4cm} >{\raggedleft\arraybackslash}p{2.4cm} >{\raggedleft\arraybackslash}p{2.4cm} >{\raggedleft\arraybackslash}p{2.4cm} >{\raggedleft\arraybackslash}p{2.4cm}@{}}
\toprule
Model & No precedents & Standard retrieval & Reference-class restriction & Reference-template restriction & Matching-template exclusion \\
\midrule
\endhead
\bottomrule
\endlastfoot
\name{gemini-3.8-flash} & 42.94 & 70.80 & 74.16 & 90.75 & 45.24 \\
\name{deepseek-v4.1-flash} & 32.63 & 62.05 & 71.44 & 92.93 & 20.77 \\
\name{Qwen3-32B} & 0.00 & 20.61 & 45.86 & 60.52 & 2.70 \\
\name{Qwen3.5-27B} & 1.08 & 36.85 & 54.16 & 70.36 & 5.45 \\
\name{RetroStepLM-RAG} & --- & 72.80 & 72.16 & 79.33 & 56.22 \\
\end{longtable}
}

\noindent\textbf{b, \name{Test-2001-2024}.}

{\footnotesize
\setlength{\tabcolsep}{3pt}
\begin{longtable}{@{}l >{\raggedleft\arraybackslash}p{2.4cm} >{\raggedleft\arraybackslash}p{2.4cm} >{\raggedleft\arraybackslash}p{2.4cm} >{\raggedleft\arraybackslash}p{2.4cm} >{\raggedleft\arraybackslash}p{2.4cm}@{}}
\toprule
Model & No precedents & Standard retrieval & Reference-class restriction & Reference-template restriction & Matching-template exclusion \\
\midrule
\endhead
\bottomrule
\endlastfoot
\name{gemini-3.8-flash} & 27.49 & 69.33 & 73.35 & 88.34 & 37.15 \\
\name{deepseek-v4.1-flash} & 19.19 & 60.87 & 71.20 & 90.23 & 17.67 \\
\name{Qwen3-32B} & 0.05 & 22.67 & 46.74 & 58.95 & 2.79 \\
\name{Qwen3.5-27B} & 0.26 & 36.12 & 52.40 & 67.11 & 4.71 \\
\name{RetroStepLM-RAG} & --- & 72.40 & 71.77 & 78.23 & 47.10 \\
\end{longtable}
}

\noindent\textbf{c, \name{Test-2025}.}

{\footnotesize
\setlength{\tabcolsep}{3pt}
\begin{longtable}{@{}l >{\raggedleft\arraybackslash}p{2.4cm} >{\raggedleft\arraybackslash}p{2.4cm} >{\raggedleft\arraybackslash}p{2.4cm} >{\raggedleft\arraybackslash}p{2.4cm} >{\raggedleft\arraybackslash}p{2.4cm}@{}}
\toprule
Model & No precedents & Standard retrieval & Reference-class restriction & Reference-template restriction & Matching-template exclusion \\
\midrule
\endhead
\bottomrule
\endlastfoot
\name{gemini-3.8-flash} & 22.85 & 35.74 & 46.45 & 73.49 & 19.35 \\
\name{deepseek-v4.1-flash} & 15.85 & 30.70 & 44.80 & 77.15 & 8.43 \\
\name{Qwen3-32B} & 0.00 & 11.56 & 20.62 & 34.52 & 1.27 \\
\name{Qwen3.5-27B} & 0.27 & 13.52 & 25.50 & 44.33 & 1.64 \\
\name{RetroStepLM-RAG} & --- & 39.40 & 43.11 & 50.11 & 30.01 \\
\end{longtable}
}

\name{RetroStepLM-RAG}, developed in this work, received retrieved precedents during both training and inference; the general-purpose models received precedents only at inference, without task-specific parameter updates.

Reference-template restriction improved accuracy relative to standard retrieval for all five models on all three test sets.
Matching-template exclusion reduced accuracy in every model--test-set combination.
On \name{Test-2025}, the exclusion condition fell below the no-precedent reference for \name{gemini-3.8-flash} (19.35\% versus 22.85\%) and \name{deepseek-v4.1-flash} (8.43\% versus 15.85\%).
For \name{deepseek-v4.1-flash}, the exclusion condition also fell below the no-precedent reference on \name{USPTO-50K} and \name{Test-2001-2024}.

Reference-class restriction improved accuracy for all four general-purpose models on all three test sets.
Its effect on \name{RetroStepLM-RAG} differed: accuracy decreased slightly on \name{USPTO-50K} and \name{Test-2001-2024} and increased on \name{Test-2025}.
Reference-class restriction therefore did not provide a uniform improvement across the evaluated models.

\sinotesub{5.4}{Interpretation and scope}

These diagnostics show that the utility of retrieved precedents depends on their selection.
Structurally similar products can arise from different transformations, and standard retrieval does not guarantee that the selected precedents share the reference transformation.

The reference-informed conditions nevertheless have a limited interpretation.
Reference classes and templates are derived from the recorded answer and are generally unavailable for prospective prediction.
Their accuracies describe performance under privileged selection conditions rather than deployable retrieval performance.

\sinotehead{6}{General-purpose model evaluation and specialized-model comparisons}

\sinotesub{6.1}{General-purpose models and inference settings}

We evaluated 14 representative general-purpose models without task-specific parameter updates: seven API-accessed models across major providers and seven open-weight Qwen models spanning 2B to 32B parameters.
Each model was evaluated on \name{USPTO-50K}, \name{Test-2001-2024} and \name{Test-2025}, both without retrieved precedents and with precedents drawn from each of three retrieval corpora.
The evaluated models are listed in \tabref{12}.

\sitablehead{12}{General-purpose models.}

\begin{longtable}{@{}p{4.2cm} p{10.0cm}@{}}
\toprule
Evaluation group & Model identifiers \\
\midrule
\endhead
\bottomrule
\endlastfoot
API-accessed models & \name{glm-5.2}; \name{qwen3.8-MAX}; \name{kimi-k3}; \name{deepseek-v4-pro}; \name{deepseek-v4.1-flash}; \name{gpt-5.6-sol}; \name{gemini-3.8-flash} \\
Open-weight Qwen models & \name{Qwen3-4B}; \name{Qwen3-8B}; \name{Qwen3-32B}; \name{Qwen3.5-2B}; \name{Qwen3.5-4B}; \name{Qwen3.5-9B}; \name{Qwen3.5-27B} \\
\end{longtable}

Models used structured sequence-to-sequence prompting to generate one final precursor-set prediction per target.
Retrieved conditions supplied five reaction precedents using the representation and structured prompt family described in \noteref{3}.

API requests used the settings in \tabref{13}.

\sitablehead{13}{API request settings.}

\begin{longtable}{@{}p{5.3cm} p{9.7cm}@{}}
\toprule
Item & Setting \\
\midrule
\endhead
\bottomrule
\endlastfoot
Temperature & 0 \\
Top-p & 1.0 \\
Maximum output tokens & 65,535 \\
Thinking & Provider-default settings \\
Incomplete-response handling & Rerun the API request until a complete response was obtained \\
Scored output & One final precursor set per target \\
Reported metric & Top-1 complete precursor-set exact match \\
\end{longtable}

Only the final precursor set was scored.
Predicted and reference molecules were standardized and compared using the common procedure in \secref{methods}.
Missing predictions, invalid molecular strings and partial precursor-set matches were counted as incorrect.

\sinotesub{6.2}{Retrieval corpora and no-precedent condition}

Each model was evaluated under four conditions, summarized in \tabref{14}.

\sitablehead{14}{Retrieval-resource conditions.}

{\small
\begin{longtable}{@{}>{\raggedright\arraybackslash}p{2.7cm} >{\raggedright\arraybackslash}p{6.0cm} r r@{}}
\toprule
Condition & Retrieval record set & Reactions & Precedents per query \\
\midrule
\endhead
\bottomrule
\endlastfoot
NoPrecedents & None & 0 & 0 \\
\name{RXNGraphormer} corpus & \texttt{RXNGraphormer-RetrievalCorpus}, constructed from the \name{RXNGraphormer} training corpus & 6,591,946 & 5 \\
USPTO corpus & \texttt{USPTO-RetrievalCorpus}, assembled by merging the training sets of \name{USPTO-50K}, \name{USPTO-MIT} and \name{USPTO-Full} & 1,217,980 & 5 \\
DianShi corpus & \texttt{DianShi-PatentRxnStep-Train-Full} & 2,712,259 & 5 \\
\end{longtable}
}

Reaction counts for the \name{RXNGraphormer} and USPTO retrieval corpora are given after deduplication.
The three retrieval corpora differ in source, scale, temporal coverage and curation; the comparison therefore evaluates retrieval sources as a whole.

\sinotesub{6.3}{General-purpose model results}

\tabref{15} reports Top-1 complete precursor-set exact-match accuracy.
Column headings identify the retrieval conditions defined in \notesec{6.2}.

\sitablehead{15}{General-purpose model accuracy across retrieval conditions (\%).}

\noindent\textbf{a, \name{USPTO-50K}.}

{\small
\begin{longtable}{@{}l r r r r@{}}
\toprule
Model & NoPrecedents & \name{RXNGraphormer} corpus & USPTO corpus & DianShi corpus \\
\midrule
\endhead
\bottomrule
\endlastfoot
\name{glm-5.2} & 20.17 & 38.89 & 46.24 & 53.45 \\
\name{qwen3.8-MAX} & 21.69 & 37.55 & 44.70 & 53.43 \\
\name{kimi-k3} & 34.39 & 49.83 & 54.70 & 59.94 \\
\name{deepseek-v4-pro} & 29.38 & 35.85 & 49.97 & 60.14 \\
\name{deepseek-v4.1-flash} & 32.63 & 41.92 & 53.33 & 62.05 \\
\name{gpt-5.6-sol} & 41.42 & 42.34 & 54.54 & 63.89 \\
\name{gemini-3.8-flash} & 42.94 & 49.99 & 63.45 & 70.80 \\
\name{Qwen3-4B} & 0.00 & 7.29 & 8.61 & 13.32 \\
\name{Qwen3-8B} & 0.14 & 9.89 & 10.63 & 11.60 \\
\name{Qwen3-32B} & 0.00 & 15.74 & 17.70 & 20.61 \\
\name{Qwen3.5-2B} & 0.00 & 2.80 & 2.48 & 3.50 \\
\name{Qwen3.5-4B} & 0.06 & 10.43 & 11.58 & 17.38 \\
\name{Qwen3.5-9B} & 0.16 & 12.68 & 13.82 & 20.87 \\
\name{Qwen3.5-27B} & 1.08 & 23.47 & 29.86 & 36.85 \\
\end{longtable}
}

\noindent\textbf{b, \name{Test-2001-2024}.}

{\small
\begin{longtable}{@{}l r r r r@{}}
\toprule
Model & NoPrecedents & \name{RXNGraphormer} corpus & USPTO corpus & DianShi corpus \\
\midrule
\endhead
\bottomrule
\endlastfoot
\name{glm-5.2} & 9.98 & 20.64 & 22.22 & 52.79 \\
\name{qwen3.8-MAX} & 10.80 & 20.06 & 22.12 & 52.87 \\
\name{kimi-k3} & 17.98 & 27.96 & 28.36 & 58.58 \\
\name{deepseek-v4-pro} & 16.03 & 19.48 & 25.28 & 59.69 \\
\name{deepseek-v4.1-flash} & 19.19 & 22.56 & 27.80 & 60.87 \\
\name{gpt-5.6-sol} & 25.09 & 24.54 & 29.91 & 63.95 \\
\name{gemini-3.8-flash} & 27.49 & 30.46 & 36.89 & 69.33 \\
\name{Qwen3-4B} & 0.00 & 2.55 & 3.08 & 14.53 \\
\name{Qwen3-8B} & 0.03 & 3.58 & 3.42 & 12.43 \\
\name{Qwen3-32B} & 0.05 & 7.03 & 7.14 & 22.67 \\
\name{Qwen3.5-2B} & 0.00 & 1.08 & 1.05 & 3.69 \\
\name{Qwen3.5-4B} & 0.00 & 4.03 & 4.00 & 18.11 \\
\name{Qwen3.5-9B} & 0.05 & 4.48 & 4.82 & 20.67 \\
\name{Qwen3.5-27B} & 0.26 & 10.56 & 12.64 & 36.12 \\
\end{longtable}
}

\noindent\textbf{c, \name{Test-2025}.}

{\small
\begin{longtable}{@{}l r r r r@{}}
\toprule
Model & NoPrecedents & \name{RXNGraphormer} corpus & USPTO corpus & DianShi corpus \\
\midrule
\endhead
\bottomrule
\endlastfoot
\name{glm-5.2} & 8.17 & 9.17 & 11.03 & 21.84 \\
\name{qwen3.8-MAX} & 8.75 & 9.76 & 10.82 & 21.58 \\
\name{kimi-k3} & 15.48 & 15.32 & 17.66 & 27.09 \\
\name{deepseek-v4-pro} & 13.20 & 10.13 & 15.06 & 29.90 \\
\name{deepseek-v4.1-flash} & 15.85 & 11.98 & 15.32 & 30.70 \\
\name{gpt-5.6-sol} & 19.19 & 11.93 & 15.91 & 30.12 \\
\name{gemini-3.8-flash} & 22.85 & 18.12 & 24.13 & 35.74 \\
\name{Qwen3-4B} & 0.00 & 1.48 & 2.17 & 8.48 \\
\name{Qwen3-8B} & 0.00 & 1.70 & 1.75 & 7.16 \\
\name{Qwen3-32B} & 0.00 & 3.39 & 3.87 & 11.56 \\
\name{Qwen3.5-2B} & 0.00 & 1.01 & 1.01 & 4.03 \\
\name{Qwen3.5-4B} & 0.05 & 1.91 & 2.01 & 7.26 \\
\name{Qwen3.5-9B} & 0.00 & 2.86 & 2.97 & 8.85 \\
\name{Qwen3.5-27B} & 0.27 & 5.67 & 6.26 & 13.52 \\
\end{longtable}
}

DianShi retrieval produced the highest accuracy among the four evaluated conditions in all 42 model--test-set combinations.
It also improved every model relative to its no-precedent condition on each test set.
Improvements from the external retrieval corpora were less uniform.
For example, on \name{Test-2025}, \name{gemini-3.8-flash} achieved 18.12\% with \name{RXNGraphormer} retrieval compared with 22.85\% without precedents, whereas DianShi retrieval increased accuracy to 35.74\%.

The strongest general-purpose model was \name{gemini-3.8-flash} with DianShi retrieval, achieving 70.80\%, 69.33\% and 35.74\% on \name{USPTO-50K}, \name{Test-2001-2024} and \name{Test-2025}, respectively.

\sinotesub{6.4}{Specialized-model evaluation protocol}

{\sloppy
We evaluated eight representative released specialized models: \name{LocalRetro}\citep{chen2021localretro}, \name{Graph2Edits}\citep{zhong2023graph2edits}, \name{R-SMILES}\citep{zhong2022rsmiles}, \name{EditRetro}\citep{han2024editretro}, \name{NAG2G}\citep{yao2024nag2g}, \name{ReactSeq}\citep{xiong2025reactseq}, \name{RXNGraphormer}\citep{xu2025rxngraphormer} and \name{RSGPT}\citep{deng2025rsgpt}.\par
}
The evaluated models span template-based (\name{LocalRetro}), graph-editing (\name{Graph2Edits}, \name{NAG2G}), sequence-based (\name{R-SMILES}, \name{EditRetro}, \name{ReactSeq}) and pretrained generative (\name{RXNGraphormer}, \name{RSGPT}) approaches.
These models were used without retraining on DianShi data.
Inputs were formatted according to each method's requirements.
Inference on \name{Test-2001-2024} and \name{Test-2025} used no molecular test-time augmentation.

Ranked predictions were evaluated using the common complete precursor-set matching procedure in \secref{methods}.
Top-$k$ accuracy measured whether at least one of the first $k$ ranked candidates matched the complete standardized reference precursor set.
Candidates were assessed in their generated rank order.
The evaluation protocol is summarized in \tabref{16}.

\sitablehead{16}{Released specialized-model evaluation protocol.}

\begin{longtable}{@{}p{6.4cm} p{7.8cm}@{}}
\toprule
Component & Protocol \\
\midrule
\endhead
\bottomrule
\endlastfoot
Model parameters & Released weights \\
Retraining on DianShi data & None \\
Input formatting & According to each method's requirements \\
Molecular test-time augmentation on \name{Test-2001-2024} and \name{Test-2025} & None \\
Scoring & Common standardized complete precursor-set exact match \\
Reported ranks & Top-1, Top-3, Top-5 and Top-10 \\
Training-data and compute budget & Not matched across models \\
\end{longtable}

\sinotesub{6.5}{Complete single-step comparison}

\tabref{17} reports complete Top-$k$ results for the eight representative released specialized models and the two DianShi models trained on \name{Train-Full}.
\name{RetroStepLM-Direct} received the target product alone; \name{RetroStepLM-RAG} additionally received five precedents retrieved from the full DianShi retrieval corpus.
Training and decoding procedures for the DianShi models are described in \noteref{3}.

The strongest general-purpose model is included as a Top-1 reference.
Dashes in its Top-3, Top-5 and Top-10 columns indicate that these ranks were not evaluated because only one final precursor set was scored per target.
Additional \name{USPTO-50K} adaptation experiments are reported separately in \noteref{8}.

\sitablehead{17}{Complete single-step performance (\%).}

\noindent\textbf{a, \name{USPTO-50K}.}

{\small
\begin{longtable}{@{}l r r r r@{}}
\toprule
Model & Top-1 & Top-3 & Top-5 & Top-10 \\
\midrule
\endhead
\bottomrule
\endlastfoot
\name{LocalRetro} & 53.4 & 77.5 & 85.9 & 92.4 \\
\name{Graph2Edits} & 56.3 & 79.2 & 86.2 & 91.0 \\
\name{R-SMILES} & 55.1 & 77.3 & 83.4 & 89.4 \\
\name{EditRetro} & 60.8 & 80.6 & 86.0 & 90.3 \\
\name{NAG2G} & 55.1 & 76.9 & 83.4 & 89.9 \\
\name{ReactSeq} & 58.9 & 80.5 & 86.4 & 91.4 \\
\name{RXNGraphormer} & 51.0 & 69.0 & 74.2 & 79.2 \\
\name{RSGPT} & 63.4 & 84.2 & 89.2 & 93.0 \\
\name{gemini-3.8-flash} + DianShi retrieval & 70.80 & --- & --- & --- \\
\name{DianShi-RetroStepLM-Direct} & 64.3 & 80.3 & 85.0 & 89.2 \\
\name{DianShi-RetroStepLM-RAG} & 72.8 & 88.6 & 92.2 & 94.9 \\
\end{longtable}
}

\noindent\textbf{b, \name{Test-2001-2024}.}

{\small
\begin{longtable}{@{}l r r r r@{}}
\toprule
Model & Top-1 & Top-3 & Top-5 & Top-10 \\
\midrule
\endhead
\bottomrule
\endlastfoot
\name{LocalRetro} & 22.7 & 35.4 & 40.8 & 45.7 \\
\name{Graph2Edits} & 22.1 & 33.9 & 38.8 & 42.7 \\
\name{R-SMILES} & 23.8 & 36.0 & 41.0 & 45.4 \\
\name{EditRetro} & 19.2 & 29.9 & 31.8 & 32.2 \\
\name{NAG2G} & 23.2 & 36.0 & 40.7 & 45.9 \\
\name{ReactSeq} & 22.8 & 35.4 & 39.7 & 43.9 \\
\name{RXNGraphormer} & 19.4 & 26.9 & 28.9 & 29.7 \\
\name{RSGPT} & 15.1 & 24.1 & 28.1 & 32.0 \\
\name{gemini-3.8-flash} + DianShi retrieval & 69.33 & --- & --- & --- \\
\name{DianShi-RetroStepLM-Direct} & 62.0 & 76.7 & 80.9 & 84.3 \\
\name{DianShi-RetroStepLM-RAG} & 72.4 & 86.4 & 89.8 & 92.7 \\
\end{longtable}
}

\noindent\textbf{c, \name{Test-2025}.}

{\small
\begin{longtable}{@{}l r r r r@{}}
\toprule
Model & Top-1 & Top-3 & Top-5 & Top-10 \\
\midrule
\endhead
\bottomrule
\endlastfoot
\name{LocalRetro} & 18.3 & 29.1 & 33.7 & 38.0 \\
\name{Graph2Edits} & 16.4 & 26.9 & 31.3 & 36.4 \\
\name{R-SMILES} & 17.0 & 28.3 & 32.8 & 37.6 \\
\name{EditRetro} & 14.8 & 22.5 & 23.8 & 24.0 \\
\name{NAG2G} & 16.4 & 27.8 & 33.1 & 38.0 \\
\name{ReactSeq} & 17.4 & 27.6 & 32.3 & 37.0 \\
\name{RXNGraphormer} & 12.5 & 18.8 & 20.3 & 21.4 \\
\name{RSGPT} & 9.7 & 15.5 & 18.3 & 22.3 \\
\name{gemini-3.8-flash} + DianShi retrieval & 35.74 & --- & --- & --- \\
\name{DianShi-RetroStepLM-Direct} & 36.4 & 53.6 & 59.4 & 64.7 \\
\name{DianShi-RetroStepLM-RAG} & 39.4 & 59.1 & 67.0 & 74.9 \\
\end{longtable}
}

On each test set, \name{gemini-3.8-flash} with DianShi retrieval exceeded the Top-1 accuracy of all eight representative released specialized models evaluated here.
Full-corpus \name{RetroStepLM-RAG} achieved the highest Top-1 accuracy among the predictors reported in this note, combining retrieval-conditioned fine-tuning with inference-time reaction precedents.

\sinotehead{7}{Multistep benchmarks and planning evaluation}

\sinotesub{7.1}{Multistep benchmarks}

Multistep evaluation used three benchmarks: \name{Pistachio Hard}\citep{yu2024doubleended}, \name{USPTO-190}\citep{chen2020retrostar} and \name{PatMol-150}.
\name{Pistachio Hard} and \name{USPTO-190} are public synthesis-planning benchmarks.
\name{PatMol-150} was constructed for this study from 2025 public patents in \name{DianShi-RxnDB}.
Benchmark composition is summarized in \tabref{18}.

\sitablehead{18}{Multistep benchmark composition.}

{\small
\setlength{\tabcolsep}{3pt}
\begin{longtable}{@{}l l r r@{}}
\toprule
Benchmark & Source & Targets & Average steps \\
\midrule
\endhead
\bottomrule
\multicolumn{4}{@{}l}{\footnotesize The minimum reference-route length for \name{PatMol-150} is 7 steps.} \\
\endlastfoot
    \name{Pistachio Hard} & Public benchmark derived from Pistachio reaction data & 100 & 7.2\citep{yu2024doubleended} \\
\name{USPTO-190} & Public benchmark introduced with \name{Retro*} & 190 & 6.0\citep{yu2024doubleended} \\
\name{PatMol-150} & This study; 2025 public patents & 150 & 8.17 \\
\end{longtable}
}

\name{Pistachio Hard} was evaluated without its original specified-starting-material constraint.
All three benchmarks were evaluated against the common starting-material inventory described in \notesec{7.3}.

\sinotesub{7.2}{\name{PatMol-150} construction}

\name{PatMol-150} was constructed through single-step reaction cleaning, route-tree assembly, applicant-source screening and final target selection.

\textbf{Single-step reaction extraction and cleaning.}
Construction started from 1,300,500 single-step reactions extracted from 2025 public patents.
Records were retained only when their SMILES parsed successfully.
Filtering removed reactions with repeated molecules within either side, overlap between reactant and product sets, multiple products, product atom mappings absent from the reactants, more than four reactant components or purely inorganic products.
Within each patent, duplicate reactions differing only in component order were merged.
These steps retained 846,230 reactions.
Extraction of local reaction templates, RDChiral\citep{coley2019rdchiral} representations and synthons was then performed, and records with unsuccessful extraction were discarded, leaving 835,875 reactions.

\textbf{Route-tree assembly.}
Following the PaRoutes approach\citep{genheden2022paroutes}, reactions within the same patent Example, or title group, were connected when the product of one reaction appeared as a reactant in another.
AND--OR synthesis trees were constructed, cyclic paths were removed, and trees whose root molecule had already appeared in pre-2025 patent data were discarded.
Cross-patent connections were subsequently made, and routes were graded by reaction completeness and template commonness.
This route-construction stage yielded 24,760 routes covering 2,923 patents.

\textbf{Applicant-source screening.}
Candidate targets were deduplicated internally and against products in \name{Train-Full}.
Applicant screening used the U.S. News 2025--2026 Best Global Universities top 100 for universities and the 2026 Fortune Global 500 pharmaceutical list and its intersection with the Fortune Global 500 healthcare list for pharmaceutical and healthcare organisations.
Applicant identities were checked using Google Patents.

\textbf{Final target selection.}
For each patent, routes leading to the final target molecule were retained.
Route length was defined as the maximum reaction depth from the target to a leaf node, $d_{\max}$, with a minimum of seven steps.
Targets were standardized and deduplicated, then excluded if they appeared among products in \name{Train-Full}, targets in \name{USPTO-190} or \name{Pistachio Hard}, or molecules in the starting-material inventory.

The final benchmark contained 150 target molecules and 150 patent reference routes.
Reference-route depths ranged from 7 to 13 steps, with a mean of 8.17 and a median of 8.
Forty-one routes, or 27.3\%, had depths of at least nine steps.
The depth distribution is reported in \tabref{19}.

\sitablehead{19}{\name{PatMol-150} reference-route depth distribution.}

\begin{longtable}{@{}l r r@{}}
\toprule
Maximum reaction depth & Routes & Percentage (\%) \\
\midrule
\endhead
\bottomrule
\endlastfoot
7 & 63 & 42.0 \\
8 & 46 & 30.7 \\
9 & 17 & 11.3 \\
10 & 9 & 6.0 \\
11 & 9 & 6.0 \\
12 & 3 & 2.0 \\
13 & 3 & 2.0 \\
Total & 150 & 100.0 \\
\end{longtable}

\sinotesub{7.3}{Starting-material inventory and planning frameworks}

All evaluations used a common RDKit-standardized inventory of 23,081,629 commercially available eMolecules building blocks\citep{emolecules}.
Inventory membership defined terminal precursor availability during search.

We evaluated \name{RetroAgent}\citep{zhu2026retroagent}, \name{Retro-R1-modified}\citep{liu2025retror1} and \name{DianShi-RetroPlanSystem-v0}.
All three used \name{Qwen3.5-397B} as the planner model.
Within each framework, the planner, inventory, Top-10 single-step candidate setting and search protocol were fixed across single-step model comparisons.

\textbf{\name{RetroAgent}.}
Evaluations used the \name{RetroAgent} planning framework with its default single-step model or one of the alternative single-step predictors listed in \notesec{7.6}.
The planner model and framework-specific search protocol remained fixed across these comparisons.

\textbf{\name{Retro-R1-modified}.}
The original inference protocol was adapted for execution by a general-purpose planner.
The system prompt explicitly specified tool syntax and two action types: invoking the single-step model and selecting a candidate reaction.
It also provided reaction-selection and search guidance.
The complete interaction history was retained, and format reminders were issued when the planner produced no tool call.
Following branch failure, the environment traversed untried sibling candidates without requiring an explicit planner-issued backtracking action.
This adapted protocol was fixed across single-step model comparisons.

\textbf{\name{DianShi-RetroPlanSystem-v0}.}
Search was represented as an AND--OR tree.
A molecule node closed when at least one candidate reaction closed; a reaction node closed only when all required precursor molecules were available in the inventory or solved through further expansion.
The planner selected an expandable molecule, and a deterministic scheduler validated actions, invoked the single-step predictor, inserted candidates, checked inventory membership and propagated closure or failure.
Invalid molecular strings and duplicate precursor sets were removed.
Candidates containing the current molecule or an ancestor were blocked from cyclic expansion.

\sinotesub{7.4}{Reaction validation and fallback}

The generative single-step models (\name{RetroStepLM-Direct/RetroStepLM-RAG}, \name{RSGPT} and \name{RXNGraphormer}) were filtered with a common \name{AOT*-style} template-based reaction-validation procedure\citep{song2025aotstar} after single-step prediction.
Candidate reactions were checked using each model's corresponding template library, with template-based rewriting where applicable.

Templates were preselected using product PatternFP fingerprints and a Tversky similarity threshold of 1 ($\alpha=0$, $\beta=1$).
Product- and reactant-side SMARTS were required to match one-to-one.
In our adaptation, we applied up to 256 matching templates with rdchiral to rewrite each candidate.
If these applications yielded no acceptable candidate, the procedure considered the 1,000 highest reactant-side fingerprint matches among the product-preselected templates, ranked the resulting reactant sets with the \name{AOT*-style} reward, and retained the best; the unsolved penalty was applied only on this fallback path, with $\gamma=0.1$.
Rewritten outputs in which the target product, or any of its components, appeared as a reactant were discarded.
These settings are summarized in \tabref{20}.

\sitablehead{20}{Reaction-validation settings.}

\begin{longtable}{@{}p{5.6cm} p{8.8cm}@{}}
\toprule
Component & Setting \\
\midrule
\endhead
\bottomrule
\endlastfoot
Template preselection & Product PatternFP; Tversky similarity 1 ($\alpha=0$, $\beta=1$) \\
Structural matching & One-to-one SMARTS matching on product and reactant sides \\
Initial template application & Up to 256 matching templates rewritten with rdchiral \\
Template-search fallback & 1,000 highest reactant-side fingerprint matches among product-preselected templates \\
Fallback ranking & \name{AOT*-style} reward on reactant sets; best retained; $\gamma=0.1$ \\
Unsolved penalty & Applied only on the template-search fallback path \\
Identity-output handling & Discard outputs in which the target product, or any of its components, appears as a reactant \\
\end{longtable}

The validation procedure was shared, whereas template libraries depended on the single-step model.
Template-library sources are listed in \tabref{21}.

\sitablehead{21}{Template libraries used in planning.}

\begin{longtable}{@{}p{6.4cm} p{3.7cm} r@{}}
\toprule
Single-step model & Template-library source & Entries \\
\midrule
\endhead
\bottomrule
\endlastfoot
\name{RetroStepLM-Direct} and \name{RetroStepLM-RAG} & \name{Train-Full} & 412,027 \\
\name{RSGPT} and \name{RXNGraphormer} & \name{DESP} USPTO library & 270,794 \\
\name{RetroAgent} default model & \name{DESP} USPTO library & 270,794 \\
\name{Retro-R1} default model & \name{Retro*} USPTO library & 381,302 \\
\end{longtable}

Because the \name{Retro-R1} default model used a template library of similar size (381,302 versus 412,027 entries) yet solved 49 percentage points fewer \name{USPTO-190} targets within the same framework, library size alone does not account for the observed differences.

\sinotesub{7.5}{Search budgets and solve-rate calculation}

Performance was assessed as target-level solve rate, defined as the fraction of targets for which at least one complete route, with all terminal precursors in the starting-material inventory, was found within the search budget.
Multiple routes for one target counted as one success.

Search budget $N$ had a framework-specific definition (\tabref{22}).

\sitablehead{22}{Search-budget units by framework.}

\begin{longtable}{@{}p{6.0cm} p{8.2cm}@{}}
\toprule
Framework & Budget unit \\
\midrule
\endhead
\bottomrule
\endlastfoot
\name{RetroAgent} & Newly expanded distinct molecules \\
\name{Retro-R1-modified} & Single-step tool calls \\
\name{DianShi-RetroPlanSystem-v0} & Expanded molecule nodes; each expansion invoked the single-step model once \\
\end{longtable}

Results at $N=50$, $100$, $200$, $300$, $400$ and $500$ were budget levels along the same search trajectory for each target and single-step model.
They were not independent reruns or averages across repeated searches.
Equal numerical budgets did not imply equal computational cost across frameworks, so single-step model comparisons were made within each framework.

Solve rate measured computational route finding under the stated inventory, validation and search protocol.
It did not establish reaction conditions, selectivity, yield or experimental executability.

\sinotesub{7.6}{Complete planning results}

\tabref{23} reports solve rates at all six budget levels and corresponds to \edfigref{5}.
\emph{Default} denotes the respective framework's default single-step model.
No default-model result is included for \name{DianShi-RetroPlanSystem-v0}.

\sitablehead{23}{Multistep solve rates across search budgets (\%).}

\noindent\textbf{a, \name{RetroAgent}.}

{\footnotesize
\begin{longtable}{@{}l l r r r r r r@{}}
\toprule
Benchmark & Single-step model & $N=50$ & $N=100$ & $N=200$ & $N=300$ & $N=400$ & $N=500$ \\
\midrule
\endhead
\bottomrule
\endlastfoot
\name{Pistachio Hard} & Default & 56.00 & 56.00 & 56.00 & 56.00 & 56.00 & 56.00 \\
\name{Pistachio Hard} & \name{RXNGraphormer} & 47.00 & 47.00 & 47.00 & 47.00 & 47.00 & 47.00 \\
\name{Pistachio Hard} & \name{RSGPT} & 76.00 & 77.00 & 79.00 & 79.00 & 79.00 & 79.00 \\
\name{Pistachio Hard} & \name{RetroStepLM-Direct} & 91.00 & 92.00 & 92.00 & 92.00 & 92.00 & 92.00 \\
\name{Pistachio Hard} & \name{RetroStepLM-RAG} & 86.00 & 90.00 & 90.00 & 90.00 & 90.00 & 90.00 \\
\name{USPTO-190} & Default & 37.89 & 38.95 & 38.95 & 38.95 & 38.95 & 38.95 \\
\name{USPTO-190} & \name{RXNGraphormer} & 33.68 & 33.68 & 33.68 & 33.68 & 33.68 & 33.68 \\
\name{USPTO-190} & \name{RSGPT} & 65.79 & 68.95 & 69.47 & 69.47 & 69.47 & 69.47 \\
\name{USPTO-190} & \name{RetroStepLM-Direct} & 90.00 & 92.63 & 93.16 & 93.16 & 93.16 & 93.16 \\
\name{USPTO-190} & \name{RetroStepLM-RAG} & 93.68 & 95.79 & 96.84 & 96.84 & 96.84 & 96.84 \\
\name{PatMol-150} & Default & 37.33 & 37.33 & 37.33 & 37.33 & 37.33 & 37.33 \\
\name{PatMol-150} & \name{RXNGraphormer} & 33.33 & 33.33 & 33.33 & 33.33 & 33.33 & 33.33 \\
\name{PatMol-150} & \name{RSGPT} & 66.00 & 68.00 & 69.33 & 69.33 & 69.33 & 69.33 \\
\name{PatMol-150} & \name{RetroStepLM-Direct} & 85.33 & 89.33 & 90.67 & 90.67 & 90.67 & 90.67 \\
\name{PatMol-150} & \name{RetroStepLM-RAG} & 84.67 & 86.67 & 89.33 & 89.33 & 89.33 & 89.33 \\
\end{longtable}
}

\noindent\textbf{b, \name{Retro-R1-modified}.}

{\footnotesize
\begin{longtable}{@{}l l r r r r r r@{}}
\toprule
Benchmark & Single-step model & $N=50$ & $N=100$ & $N=200$ & $N=300$ & $N=400$ & $N=500$ \\
\midrule
\endhead
\bottomrule
\endlastfoot
\name{Pistachio Hard} & Default & 69.00 & 69.00 & 69.00 & 69.00 & 69.00 & 69.00 \\
\name{Pistachio Hard} & \name{RXNGraphormer} & 35.00 & 36.00 & 36.00 & 36.00 & 36.00 & 36.00 \\
\name{Pistachio Hard} & \name{RSGPT} & 78.00 & 78.00 & 79.00 & 79.00 & 79.00 & 79.00 \\
\name{Pistachio Hard} & \name{RetroStepLM-Direct} & 84.00 & 84.00 & 84.00 & 84.00 & 84.00 & 84.00 \\
\name{Pistachio Hard} & \name{RetroStepLM-RAG} & 91.00 & 92.00 & 92.00 & 92.00 & 92.00 & 92.00 \\
\name{USPTO-190} & Default & 43.16 & 43.16 & 43.16 & 43.16 & 43.16 & 43.16 \\
\name{USPTO-190} & \name{RXNGraphormer} & 28.42 & 30.53 & 30.53 & 31.05 & 31.05 & 31.05 \\
\name{USPTO-190} & \name{RSGPT} & 57.37 & 57.89 & 61.05 & 61.05 & 61.05 & 61.05 \\
\name{USPTO-190} & \name{RetroStepLM-Direct} & 83.16 & 83.68 & 85.26 & 85.26 & 85.26 & 85.26 \\
\name{USPTO-190} & \name{RetroStepLM-RAG} & 90.00 & 91.05 & 92.11 & 92.11 & 92.11 & 92.11 \\
\name{PatMol-150} & Default & 52.00 & 52.00 & 52.00 & 52.00 & 52.00 & 52.00 \\
\name{PatMol-150} & \name{RXNGraphormer} & 22.00 & 23.33 & 23.33 & 23.33 & 23.33 & 23.33 \\
\name{PatMol-150} & \name{RSGPT} & 63.33 & 65.33 & 67.33 & 68.00 & 68.00 & 68.00 \\
\name{PatMol-150} & \name{RetroStepLM-Direct} & 82.00 & 83.33 & 83.33 & 84.00 & 84.00 & 84.00 \\
\name{PatMol-150} & \name{RetroStepLM-RAG} & 83.33 & 85.33 & 86.00 & 86.00 & 86.00 & 86.00 \\
\end{longtable}
}

\noindent\textbf{c, \name{DianShi-RetroPlanSystem-v0}.}

{\footnotesize
\begin{longtable}{@{}l l r r r r r r@{}}
\toprule
Benchmark & Single-step model & $N=50$ & $N=100$ & $N=200$ & $N=300$ & $N=400$ & $N=500$ \\
\midrule
\endhead
\bottomrule
\endlastfoot
\name{Pistachio Hard} & \name{RXNGraphormer} & 39.00 & 40.00 & 44.00 & 44.00 & 44.00 & 44.00 \\
\name{Pistachio Hard} & \name{RSGPT} & 76.00 & 78.00 & 86.00 & 88.00 & 88.00 & 88.00 \\
\name{Pistachio Hard} & \name{RetroStepLM-Direct} & 88.00 & 88.00 & 90.00 & 93.00 & 94.00 & 94.00 \\
\name{Pistachio Hard} & \name{RetroStepLM-RAG} & 89.00 & 91.00 & 94.00 & 95.00 & 96.00 & 96.00 \\
\name{USPTO-190} & \name{RXNGraphormer} & 27.37 & 32.11 & 34.74 & 35.79 & 35.79 & 36.32 \\
\name{USPTO-190} & \name{RSGPT} & 61.58 & 65.79 & 81.05 & 82.63 & 83.68 & 83.68 \\
\name{USPTO-190} & \name{RetroStepLM-Direct} & 88.42 & 91.58 & 94.21 & 96.84 & 97.37 & 97.37 \\
\name{USPTO-190} & \name{RetroStepLM-RAG} & 92.11 & 94.21 & 96.32 & 96.84 & 96.84 & 96.84 \\
\name{PatMol-150} & \name{RXNGraphormer} & 24.67 & 26.67 & 26.67 & 26.67 & 26.67 & 27.33 \\
\name{PatMol-150} & \name{RSGPT} & 68.67 & 70.67 & 76.00 & 80.00 & 81.33 & 81.33 \\
\name{PatMol-150} & \name{RetroStepLM-Direct} & 86.67 & 89.33 & 94.00 & 96.00 & 97.33 & 97.33 \\
\name{PatMol-150} & \name{RetroStepLM-RAG} & 90.67 & 91.33 & 95.33 & 95.33 & 95.33 & 95.33 \\
\end{longtable}
}

\name{RetroStepLM-Direct} and \name{RetroStepLM-RAG} were developed in this work.

Both DianShi models exceeded every evaluated non-DianShi model across all nine framework--benchmark settings at both $N=50$ and $N=500$.
At $N=500$, DianShi solve rates ranged from 84.00\% to 97.37\%.
\name{RetroStepLM-Direct} solved more targets than \name{RetroStepLM-RAG} in four of the nine settings at $N=500$ and in two of the nine at $N=50$, despite \name{RetroStepLM-RAG}'s higher single-step Top-1 accuracy.

\sinotehead{8}{Additional ablations and retrieval corpus transfer}

\sinotesub{8.1}{Intermediate-label and retrieval ablations}

We evaluated combinations of intermediate-label supervision and retrieval conditioning using \name{Qwen3-4B} as the common base model.
Four models received task-specific training on \name{Train-Full}, with intermediate labels and retrieval enabled or omitted.
\name{Qwen3-4B} without task-specific training or retrieved precedents provided an additional reference.

Models with intermediate labels included the synthon and leaving-group fields described in \noteref{3} in their training targets.
Models without intermediate labels omitted these fields.
Retrieval-enabled models used precedents from \name{Train-Full} during training and inference; models without retrieval received the query product alone.
The models with intermediate labels, without and with retrieval, correspond to the full-corpus \name{RetroStepLM-Direct} and \name{RetroStepLM-RAG} models, respectively.

\tabref{24} reports complete precursor-set exact-match accuracy.
The four trained models form a comparison of intermediate-label and retrieval combinations; the untrained reference additionally differs in whether task-specific parameter updates were performed.

\sitablehead{24}{Intermediate-label and retrieval ablation results (\%).}

\noindent\textbf{a, \name{USPTO-50K}.}

{\small
\begin{longtable}{@{}l l l r r r r@{}}
\toprule
Training corpus & Intermediate labels & Retrieval & Top-1 & Top-3 & Top-5 & Top-10 \\
\midrule
\endhead
\bottomrule
\endlastfoot
None & No & No & 0.0 & 0.0 & 0.0 & 0.0 \\
\name{Train-Full} & No & No & 61.49 & 78.37 & 82.98 & 86.74 \\
\name{Train-Full} & Yes & No & 64.3 & 80.3 & 85.0 & 89.2 \\
\name{Train-Full} & No & Yes & 71.70 & 86.84 & 90.71 & 94.19 \\
\name{Train-Full} & Yes & Yes & 72.8 & 88.6 & 92.2 & 94.9 \\
\end{longtable}
}

\noindent\textbf{b, \name{Test-2001-2024}.}

{\small
\begin{longtable}{@{}l l l r r r r@{}}
\toprule
Training corpus & Intermediate labels & Retrieval & Top-1 & Top-3 & Top-5 & Top-10 \\
\midrule
\endhead
\bottomrule
\endlastfoot
None & No & No & 0.0 & 0.0 & 0.0 & 0.0 \\
\name{Train-Full} & No & No & 58.29 & 73.85 & 78.07 & 82.10 \\
\name{Train-Full} & Yes & No & 62.0 & 76.7 & 80.9 & 84.3 \\
\name{Train-Full} & No & Yes & 70.06 & 84.81 & 88.47 & 91.73 \\
\name{Train-Full} & Yes & Yes & 72.4 & 86.4 & 89.8 & 92.7 \\
\end{longtable}
}

\noindent\textbf{c, \name{Test-2025}.}

{\small
\begin{longtable}{@{}l l l r r r r@{}}
\toprule
Training corpus & Intermediate labels & Retrieval & Top-1 & Top-3 & Top-5 & Top-10 \\
\midrule
\endhead
\bottomrule
\endlastfoot
None & No & No & 0.0 & 0.0 & 0.0 & 0.0 \\
\name{Train-Full} & No & No & 35.63 & 51.48 & 57.26 & 63.26 \\
\name{Train-Full} & Yes & No & 36.4 & 53.6 & 59.4 & 64.7 \\
\name{Train-Full} & No & Yes & 34.36 & 54.19 & 61.40 & 69.19 \\
\name{Train-Full} & Yes & Yes & 39.4 & 59.1 & 67.0 & 74.9 \\
\end{longtable}
}

Including intermediate labels increased accuracy at every reported rank in both retrieved and non-retrieved trained models.
The combination of intermediate labels and retrieval achieved the highest accuracy among these models on all three test sets.
Retrieval alone did not improve every metric: without intermediate labels, \name{Test-2025} Top-1 accuracy decreased from 35.63\% to 34.36\%, although Top-3, Top-5 and Top-10 accuracy increased.

\sinotesub{8.2}{Training-time precedent sampling}

We compared two training-time precedent-selection strategies for full-corpus \name{RetroStepLM-RAG}: deterministic selection of the five highest-similarity eligible precedents, and random selection of five precedents from the ten highest-similarity eligible candidates at each epoch.
Both models used \name{Train-Full}, intermediate labels and deterministic Top-5 retrieval at inference.
The comparison changed training-time precedent selection while retaining the common training and evaluation protocol.

These were separately trained models.
Their comparison therefore differs from the fixed-checkpoint inference interventions in \noteref{4}.
Results are reported in \tabref{25}.

\sitablehead{25}{Training-time precedent-sampling results (\%).}

\noindent\textbf{a, \name{USPTO-50K}.}

{\small
\begin{longtable}{@{}l r r r r@{}}
\toprule
Training-time selection & Top-1 & Top-3 & Top-5 & Top-10 \\
\midrule
\endhead
\bottomrule
\endlastfoot
Deterministic Tanimoto Top-5 & 71.66 & 87.20 & 91.33 & 94.21 \\
Random five from Tanimoto Top-10 & 72.8 & 88.6 & 92.2 & 94.9 \\
\end{longtable}
}

\noindent\textbf{b, \name{Test-2001-2024}.}

{\small
\begin{longtable}{@{}l r r r r@{}}
\toprule
Training-time selection & Top-1 & Top-3 & Top-5 & Top-10 \\
\midrule
\endhead
\bottomrule
\endlastfoot
Deterministic Tanimoto Top-5 & 70.19 & 84.99 & 88.84 & 92.18 \\
Random five from Tanimoto Top-10 & 72.4 & 86.4 & 89.8 & 92.7 \\
\end{longtable}
}

\noindent\textbf{c, \name{Test-2025}.}

{\small
\begin{longtable}{@{}l r r r r@{}}
\toprule
Training-time selection & Top-1 & Top-3 & Top-5 & Top-10 \\
\midrule
\endhead
\bottomrule
\endlastfoot
Deterministic Tanimoto Top-5 & 36.80 & 54.88 & 62.25 & 70.15 \\
Random five from Tanimoto Top-10 & 39.4 & 59.1 & 67.0 & 74.9 \\
\end{longtable}
}

Random selection from the Top-10 candidates produced higher accuracy at every reported rank on all three test sets.
This strategy was used for the main \name{RetroStepLM-RAG} model.
The results describe the evaluated training runs and do not estimate variability across repeated training.

\sinotesub{8.3}{Additional \name{USPTO-50K} adaptation}

We additionally evaluated a two-stage \name{RetroStepLM-Direct} model.
Starting from the model trained on \name{Train-Full}, a second training stage adapted the model on \name{USPTO-50K-train}.
Both the original and adapted models predicted from the query product without retrieved precedents.

This model differs from both the unadapted full-corpus \name{RetroStepLM-Direct} model and the USPTO-only \name{RetroStepLM-RAG} model described in \notesec{8.4}.
Its training history includes DianShi training followed by \name{USPTO-50K} adaptation.
Results are reported in \tabref{26}.

\sitablehead{26}{\name{RetroStepLM-Direct} performance before and after \name{USPTO-50K} adaptation (\%).}

\noindent\textbf{a, \name{USPTO-50K}.}

{\small
\begin{longtable}{@{}l r r r r@{}}
\toprule
Training history & Top-1 & Top-3 & Top-5 & Top-10 \\
\midrule
\endhead
\bottomrule
\endlastfoot
\name{Train-Full} & 64.3 & 80.3 & 85.0 & 89.2 \\
\name{Train-Full} $\rightarrow$ \name{USPTO-50K-train} & 67.9 & 86.0 & 90.9 & 95.0 \\
\end{longtable}
}

\noindent\textbf{b, \name{Test-2001-2024}.}

{\small
\begin{longtable}{@{}l r r r r@{}}
\toprule
Training history & Top-1 & Top-3 & Top-5 & Top-10 \\
\midrule
\endhead
\bottomrule
\endlastfoot
\name{Train-Full} & 62.0 & 76.7 & 80.9 & 84.3 \\
\name{Train-Full} $\rightarrow$ \name{USPTO-50K-train} & 57.4 & 76.2 & 81.7 & 86.5 \\
\end{longtable}
}

\noindent\textbf{c, \name{Test-2025}.}

{\small
\begin{longtable}{@{}l r r r r@{}}
\toprule
Training history & Top-1 & Top-3 & Top-5 & Top-10 \\
\midrule
\endhead
\bottomrule
\endlastfoot
\name{Train-Full} & 36.4 & 53.6 & 59.4 & 64.7 \\
\name{Train-Full} $\rightarrow$ \name{USPTO-50K-train} & 34.6 & 52.9 & 60.9 & 69.5 \\
\end{longtable}
}

Adaptation increased \name{USPTO-50K} accuracy at all reported ranks.
On both DianShi test sets, Top-1 and Top-3 accuracy decreased, whereas Top-5 and Top-10 accuracy increased.
The effect of adaptation therefore depended on both the evaluation set and candidate rank.

\sinotesub{8.4}{Retrieval corpus transfer for a USPTO-only \name{RetroStepLM-RAG} model}

A separate \name{RetroStepLM-RAG} model was initialized from \name{Qwen3-4B} and trained only on \name{USPTO-50K-train} using intermediate labels and retrieval-conditioned fine-tuning.
\name{USPTO-50K-train} supplied both the supervised reactions and training precedents.
This model did not undergo task-specific training on \name{Train-Full}.

After training, the checkpoint was fixed and evaluated with three inference retrieval corpora: \name{USPTO-50K-train}, the DianShi \name{Pre2017} corpus and \name{Train-Full}.
The model's task-specific training history remained unchanged across these conditions.

\name{USPTO-50K-train} here denotes the individual benchmark's training partition.
It is distinct from \texttt{USPTO-RetrievalCorpus} in \noteref{6}, which merges the training sets of \name{USPTO-50K}, \name{USPTO-MIT} and \name{USPTO-Full}.
Results are reported in \tabref{27}.

\sitablehead{27}{Retrieval corpus transfer for the USPTO-only \name{RetroStepLM-RAG} checkpoint (\%).}

\noindent\textbf{a, \name{USPTO-50K}.}

{\small
\begin{longtable}{@{}l r r r r@{}}
\toprule
Inference retrieval corpus & Top-1 & Top-3 & Top-5 & Top-10 \\
\midrule
\endhead
\bottomrule
\endlastfoot
\name{USPTO-50K-train} & 53.88 & 73.66 & 80.27 & 85.98 \\
\name{Pre2017} & 64.81 & 81.85 & 86.90 & 91.13 \\
Full & 64.57 & 81.33 & 87.00 & 91.31 \\
\end{longtable}
}

\noindent\textbf{b, \name{Test-2001-2024}.}

{\small
\begin{longtable}{@{}l r r r r@{}}
\toprule
Inference retrieval corpus & Top-1 & Top-3 & Top-5 & Top-10 \\
\midrule
\endhead
\bottomrule
\endlastfoot
\name{USPTO-50K-train} & 20.75 & 31.49 & 35.49 & 40.44 \\
\name{Pre2017} & 37.52 & 49.87 & 53.98 & 58.48 \\
Full & 44.92 & 57.82 & 61.90 & 66.14 \\
\end{longtable}
}

\noindent\textbf{c, \name{Test-2025}.}

{\small
\begin{longtable}{@{}l r r r r@{}}
\toprule
Inference retrieval corpus & Top-1 & Top-3 & Top-5 & Top-10 \\
\midrule
\endhead
\bottomrule
\endlastfoot
\name{USPTO-50K-train} & 14.48 & 23.17 & 27.68 & 31.97 \\
\name{Pre2017} & 16.44 & 26.56 & 31.28 & 36.00 \\
Full & 21.16 & 32.24 & 37.01 & 42.05 \\
\end{longtable}
}

Both DianShi retrieval corpora improved accuracy relative to retrieval from \name{USPTO-50K-train} at every reported rank on all three test sets.
Full retrieval further improved all reported metrics on \name{Test-2001-2024} and \name{Test-2025} relative to \name{Pre2017} retrieval.
On \name{USPTO-50K}, the differences between the two DianShi corpora were small and varied by rank.

These results support the utility of updating inference-time retrieval resources when the available corpus differs from the model's task-specific training corpus.